\documentclass{article}

\usepackage{PRIMEarxiv}

\usepackage[utf8]{inputenc} 
\usepackage[T1]{fontenc}    
\usepackage{hyperref}       
\usepackage{url}            
\usepackage{booktabs}       
\usepackage{amsfonts}       
\usepackage{nicefrac}       
\usepackage{microtype}      
\usepackage{lipsum}
\usepackage{fancyhdr}       
\usepackage{graphicx}       
\graphicspath{{media/}}     
 \usepackage{amsmath}
 \usepackage{caption}

\title{Consistency Regularisation in Varying Contexts and Feature Perturbations for Semi-Supervised Semantic Segmentation of Histology Images
}

\author{
  Raja Muhammad Saad Bashir \\
  Tissue Image Analytics Centre \\
  University of Warwick\\
  Coventry, United Kingdom \\
  \texttt{saad.bashir@warwick.ac.uk} \\
   \And
   Talha Qaiser \\
  Tissue Image Analytics Centre \\
  University of Warwick\\
  Coventry, United Kingdom \\
  \texttt{talha.qaiser@warwick.ac.uk} \\
 \And
   Shan E Ahmed Raza \\
  Tissue Image Analytics Centre \\
  University of Warwick\\
  Coventry, United Kingdom \\
  \texttt{shan.raza@warwick.ac.uk} \\
\And
   Nasir M. Rajpoot \\
  Tissue Image Analytics Centre \\
  University of Warwick\\
  Coventry, United Kingdom \\
  \texttt{n.m.rajpoot@warwick.ac.uk} \\
}

\begin{document}
\maketitle

\begin{abstract}
Semantic segmentation of various tissue and nuclei types in histology images is fundamental to many downstream tasks in the area of computational pathology (CPath). In recent years, Deep Learning (DL) methods have been shown to perform well on segmentation tasks but DL methods generally require a large amount of pixel-wise annotated data. Pixel-wise annotation sometimes requires expert's knowledge and time which is laborious and costly to obtain. In this paper, we present a consistency based semi-supervised learning (SSL) approach that can help mitigate this challenge by exploiting a large amount of unlabelled data for model training thus alleviating the need for a large annotated dataset. However, SSL models might also be susceptible to changing context and features perturbations exhibiting poor generalisation due to the limited training data. We propose an SSL method that learns robust features from both labelled and unlabelled images by enforcing consistency against varying contexts and feature perturbations. The proposed method incorporates context-aware consistency by contrasting pairs of overlapping images in a pixel-wise manner from changing contexts resulting in robust and context invariant features. We show that cross-consistency training makes the encoder features invariant to different perturbations and improves the prediction confidence. Finally, entropy minimisation is employed to further boost the confidence of the final prediction maps from unlabelled data. We conduct an extensive set of experiments on two publicly available large datasets (BCSS and MoNuSeg) and show superior performance compared to the state-of-the-art methods.
\end{abstract}


\section{Introduction}
Segmentation of fundamental objects and regions in histology images is key to several downstream analysis tasks in computational pathology (CPath) \cite{koohbanani2020nuclick,shephard2021simultaneous} e.g., cancer type classification \cite{qaiser2019fast,vu2021digital,jahanifar2021stain,lu2021data}, tumour and glandular segmentation \cite{da2022digestpath}, and other tasks like mutation prediction \cite{bilal2021development,echle2022artificial}. Their utility is not limited to diagnosis, they have also been employed for prognostic purposes e.g., tumour infiltrating lymphocytes (TILs) have been found to be significant prognostic biomarker in various types of tumours \cite{shaban2022digital}. Similarly, tumour progression has been linked with interaction between the tumour epithelial cells and tumour associate stroma \cite{mao2013stromal}. Hence, it is important to segment different types of histological objects precisely as their quantification is vital to downstream analysis.

Machine learning based traditional methods accomplished this task using different hand-crafted features e.g., colour \cite{tabesh2007multifeature}, texture \cite{diamond2004use,sirinukunwattana2015novel} and morphological features \cite{anoraganingrum1999cell}. Recently, deep learning (DL) algorithms have gained increasing attention in semantic segmentation due to their superior performance on natural and medical images \cite{ronneberger2015u,Chen2017DeepLab,xie2021segformer,zhang2022topformer}.   However, DL methods are known to be ``data hungry" and require a large amount of annotated data. Precise annotation of histology images is an expensive and laborious process, requiring up to $\sim$5-6 hours of an expert histopathologist's time to annotate one whole-slide image (WSI) \cite{lindman2019annotations}. To alleviate the annotation burden, other modes of training have been proposed such as patch based segmentation \cite{shaban2019novel,kather2019deep}, coarse segmentation \cite{bejnordi2015multi,bashir2022novel} and interactive segmentation \cite{koohbanani2020nuclick,jahanifar2021robust} but these methods still require large-scale weak annotations involving human expert.

Semantic segmentation is a pixel-level classification task of predicting label for each pixel using pixel values. Most of the early DL methods were based on fully convolutional networks (FCN) \cite{long2015Fully} where pooling layers aggregate the information by focusing on ``what" rather than ``where" resulting in loss of spatial information. The subsequent studies addressed this shortcoming by using pooling layers with more advanced techniques involving skip connections, encoders and boundary in formation. As semantic segmentation is more than just assigning labels to pixels, it inevitably requires some contextual information along with knowledge of colour, edges and resolutions. In this regard, algorithms like UNet \cite{ronneberger2015u}, PSPNet \cite{zhao2017pyramid}, HRNet \cite{sun2019deep} and DeepLab-v3 \cite{Chen2017DeepLab} use techniques like encoder-decoder architecture, wider receptive fields and dilated/\textit{atrous} convolutions to improve the segmentation performance. More recently, another line of work focused on transforming the task of semantic segmentation to sequence-to-sequence prediction, where, self-attention mechanism is introduced using transformers \cite{vaswani2017attention} to encode the global context in each layer \cite{zheng2021rethinking,xie2021segformer} for subsequent decoding. However, a downside of using transformer based technique is their computational complexity. 
 
On the other hand, semi-supervised learning (SSL) can train DL models with a small set of annotated data by leveraging the unlabelled data for better representation learning hence boosting the performance. SSL methods consist of different techniques to incorporate unlabelled data for learning including pseudo labelling \cite{lee2013pseudo,zou2020pseudoseg,chen2021semi}, generative adversarial modelling \cite{goodfellow2014yoshua,badrinarayanan2015deep,kumar2017semi,miyato2018virtual}, consistency training \cite{wei2018revisiting,ouali2020semi,chen2021semi,lai2021semi} and entropy minimisation \cite{grandvalet2004semi,saito2019semi,wu2022cross}. However, SSL methods have an additional issue related to overfitting of small labelled input data which may lead to poor generalisation. 
 
\begin{figure}[!ht]
    \centering
    \includegraphics[width=0.75\textwidth]{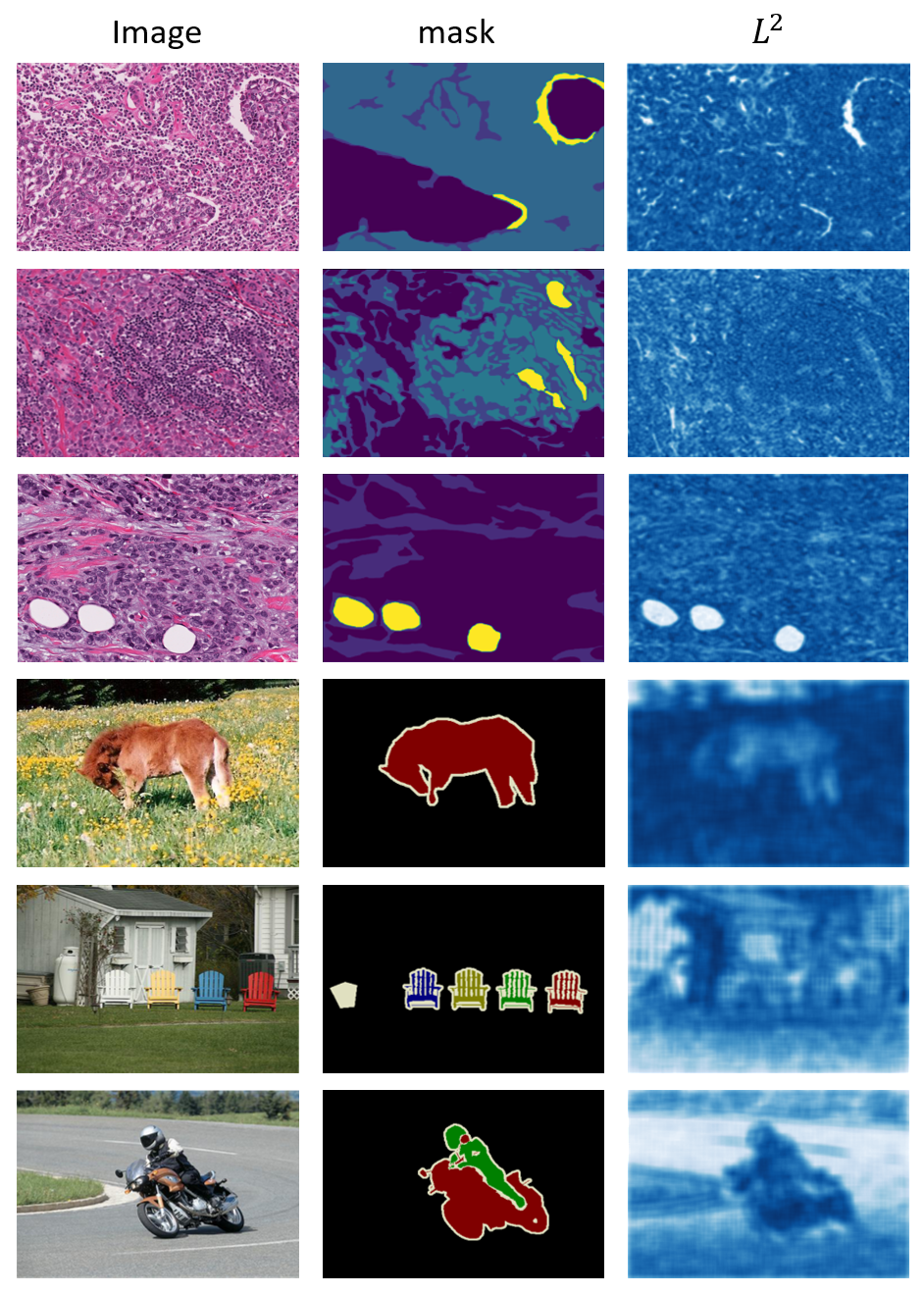}
    \caption{ ($1^{st}$ column) Example images from histological (BCSS) and natural (PASCAL VOC 2012) datasets; ($2^{nd}$ column) Respective masks showing the foreground and background objects with boundaries; ($3^{rd}$ column) Average Euclidean distance $L^2$ between the central patch of size $21 \times 21$ with four overlapping patches in the immediate neighbours in RGB colour space. Note that the darker blue colour represents the low density regions corresponding to high average distance. }
    \label{fig:cluster_assumption}
\end{figure}

\begin{figure}
    \centering
    \includegraphics[width=0.75\textwidth]{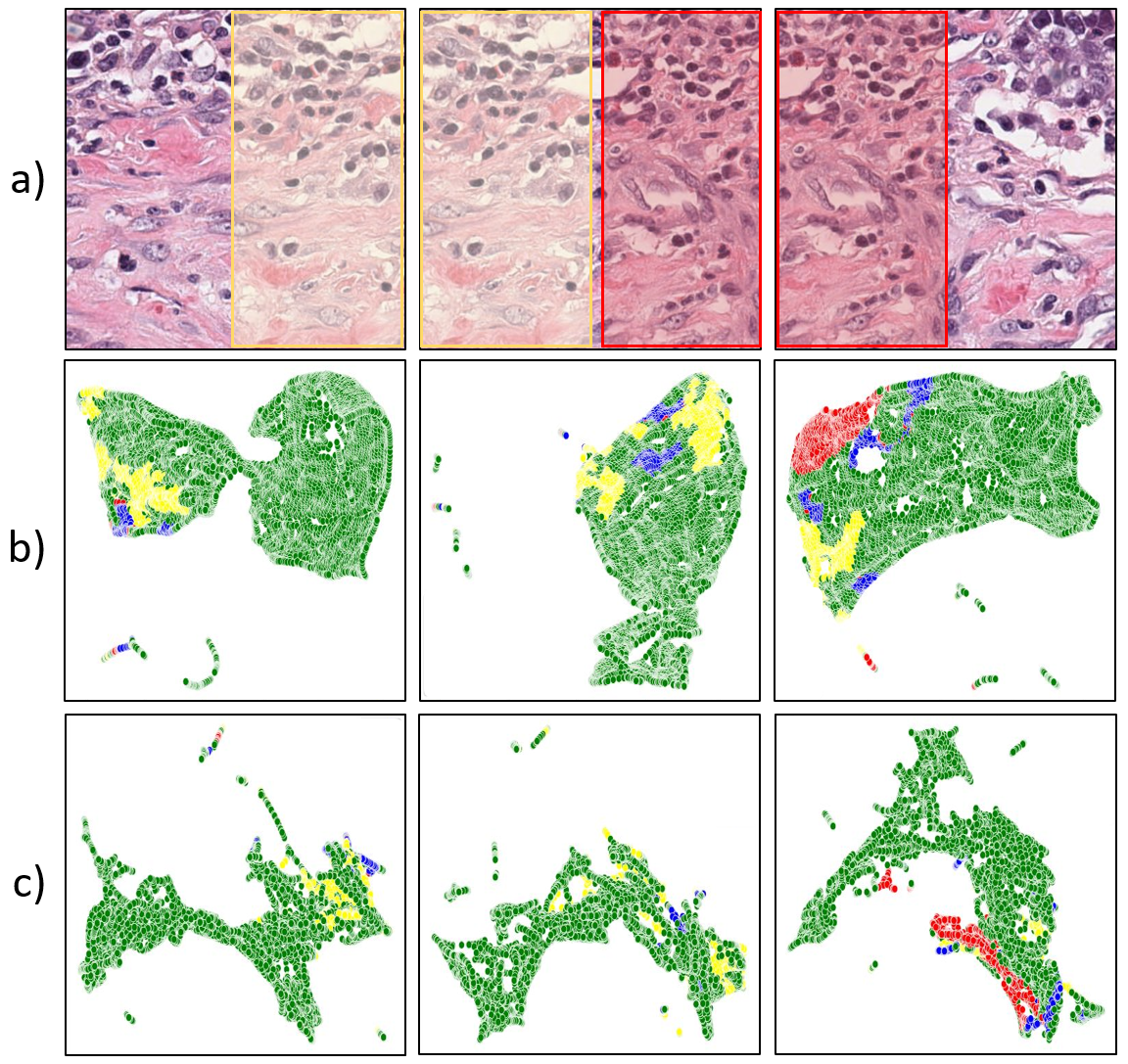}
    \caption{(a) Images from the BCSS dataset with overlapping regions cropped sequentially from the same image to mimic changing contexts; (b) UMAP visualisations of features embedding distributions extracted from a fully supervised model; (c) UMAP visualisations of feature embedding distributions extracted from our semi-supervised model. Note that the feature embeddings are represented in the same UMAP space where dots with same colour represents feature embedding from the same class.}
    \label{fig:intro_context_umap}
\end{figure}

In this paper, we propose a novel consistency based SSL method  for semantic segmentation which leverages unlabelled data in varying contexts and feature perturbations. Consistency regularisation is enforced by using context-aware contrastive learning in changing contexts and cross-consistency training is used to handle feature perturbations along with entropy minimisation for confident predictions. The main purpose of consistency regularisation is to enforce the model to output consistent predictions for unlabelled data under changing conditions. For consistency to work effectively, input space must hold the cluster assumption constraint i.e., same label is most likely to be shared among the nearby samples thus forming a cluster. Therefore, high density regions would correspond to clusters (i.e., samples with same labels) whereas the low density regions are separation spaces (i.e., object boundaries). As for histology and natural images, the pixel space might not hold the constraint of cluster assumption as it can be seen in Figure \ref{fig:cluster_assumption}. The low density regions (i.e., high average distance) do not align well with the class boundaries in most of the scenarios e.g., in $1^{st}$ row we observe low density regions throughout the image, while, in the last row there exist a cluster of high density regions for foreground object i.e., road. However, the cluster assumption holds in encoders latent feature space \cite{ouali2020semi}, as we show and discuss later in this paper's Figure \ref{fig:encoder_cluster_assumtion}. Therefore, we applied the feature perturbations to encoders output rather than the input images. Also, due to the limited labelled data, the model may become overly dependent on just context overlooking the objects themselves, losing self-awareness \cite{lai2021semi}. Therefore, to enforce consistency against changing contexts, we propose context-aware contrastive learning which helps the model learn high-level semantic features by contrasting the positive and negative pairs of images in different contexts. As shown in Figure \ref{fig:intro_context_umap}, under varying contexts the model trained in a fully supervised manner is unable to produce consistent feature distributions as compared to our proposed method consistency regularisation in varying contexts and feature perturbations for semi-supervised semantic segmentation of Histology Images (CRCFP) with consistent feature distributions. While context-aware consistency brings robustness to changing contexts, cross-consistency training can help the model learn invariant feature representations that is robust to small perturbations. While context-aware and cross-consistency training regularisation can bring consistency in encoder's features representations, it often fails to optimise the pixel classifier leading to less confident prediction maps. Finally, entropy minimisation coupled with aforementioned techniques helps the model acquire high quality and confident predictions. We extensively evaluated our proposed CRCFP on two publicly available histology image datasets BCSS \cite{amgad2019structured} and MoNuSeg \cite{kumar2019multi} for two different semantic segmentation tasks i.e., tissue region segmentation and nuclei segmentation. In summary, our contributions are as follows:

\begin{itemize}
    \item We propose a consistency regularisation based SSL method against varying contexts and perturbations using a novel combination of context-aware consistency loss and cross-consistent training for feature generalisability. 
    \item To improve the confidence of final predictions for pseudo labelling, entropy minimisation is employed on top of context-aware and cross-consistent regularisation.
    \item We demonstrate our method on two different semantic segmentation tasks i.e., cancer region and nuclei segmentation on two publicly available large histology datasets.
    \item Extensive experiments showed superior performance of our method outperforming the state-of-the-art (SOTA) SSL methods with extensive ablation studies.
\end{itemize}

\section{Literature Review}
\subsection{Semantic Segmentation} The transformation of pixel values of an image to class labels using high level features is known as semantic segmentation and is fundamentally a challenging task. FCN extracts meaningful visual hierarchical features for various computer vision tasks e.g., classification, segmentation and object detection. However, due to the pooling layers spatial information is lost in aggregation which is vital in segmentation tasks and results in smaller output \cite{long2015Fully}. Encoder-decoder based architectures solve this issue by recovering and refining the output spatially in a step wise fashion \cite{Badrinarayanan2016SegNet,wang2016deep,naylor2017nuclei}. Further improvements can be made possible with the help of skip connections which results in more refined boundaries and confident predictions \cite{ronneberger2015u}. However, the downside of the encoder-decoder architectures is a limited receptive field resulting in missing long-range dependencies. Dilated/atrous convolutions \cite{Chen2017DeepLab,yu2015multi,azam2020novel,bashir2022novel}, spatial pyramid pooling \cite{he2015spatial,zhao2017pyramid,graham2019mild,da2022digestpath} and attention based algorithms \cite{zhu2019asymmetric,huang2019ccnet,fraz2020fabnet,xie2021segformer} can enable aggregation of context by using larger receptive fields or maintaining spatial information. More recently attention mechanism \cite{vaswani2017attention} has been used to replace limited local receptive field of convolutions with global contexts using transformers. Images are transformed into sequence of patches for transformer \cite{dosovitskiy2020image} to process as transformers capture more consistent global contexts due to their self-attention mechanisms \cite{zheng2021rethinking,xie2021segformer,karimi2022medical}. Despite the advancements and improvements in semantic segmentation the bottleneck for high accuracy still remains to be dependent on pixel-wise annotations.

\subsection{Semi-Supervised Learning} Semi-supervised learning (SSL) exploits the unlabelled data on top of limited labelled data for improving the model performance and internal feature representation. Recently SSL based methods have been widely adopted in the computer vision domain \cite{van2020survey}. Popular SSL techniques include pseudo labelling \cite{lee2013pseudo,zou2020pseudoseg,chen2021semi} where the model trained on limited data is used to predict the labels for unlabelled data known as pseudo labels. Generative adversarial based methods improve the generalisability of the trained model using various perturbations in the direction of maximum vulnerability, resulting in aligning the distributions of labelled and unlabelled input in latent space \cite{badrinarayanan2015deep,kumar2017semi,wei2017object,miyato2018virtual}. Data interpolation based methods aim to augment input space to create perturbed linear inputs for training models \cite{berthelot2019mixmatch,berthelot2019remixmatch,bashir2020hydramix}, Temporal ensembling based methods aim to ensemble predictions over the epochs using momentum/moving average to enforce consistency between the predictions \cite{laine2016temporal,tarvainen2017mean}. Self-supervised learning based consistency training aims to contrast the unlabelled input using pre-text tasks for learning important representations \cite{chen2020simple,he2020momentum,koohbanani2021self, chen2021semi,lai2021semi} and entropy minimisation based method aims to maximise label assignment to either of the labels \cite{grandvalet2004semi,saito2019semi,wu2022cross}. 

\subsection{Contrastive Learning}
Learning by contrasting pairs of similar (positive) and dissimilar (negative) images for improved representation learning is known as contrastive learning \cite{hadsell2006dimensionality,chen2020simple,liu2021domain}. Several loss functions have been proposed from maximum margin loss \cite{chopra2005learning}, triplet loss \cite{schroff2015facenet}, N-pair loss \cite{sohn2016improved} to contrastive predicting coding (CPC) \cite{oord2018representation} proposing mutual information based InfoNCE loss to improve contrastive learning. Contrastive learning has been used in both supervised and unsupervised learning tasks in conjunction with self supervision \cite{he2020momentum,chen2020simple,grill2020bootstrap}.  Recently, it has been established that using more accurate positive and negative pairs along with larger batch sizes improves the quality of learned representations with heavy augmentations. Memory banks are adopted when large batches are not computationally feasible (i.e., doesn't fit the GPU memory) for contrastive loss using a large set of negative samples. 

\begin{figure}[!t]
    \centering
    \includegraphics[width=0.9\textwidth]{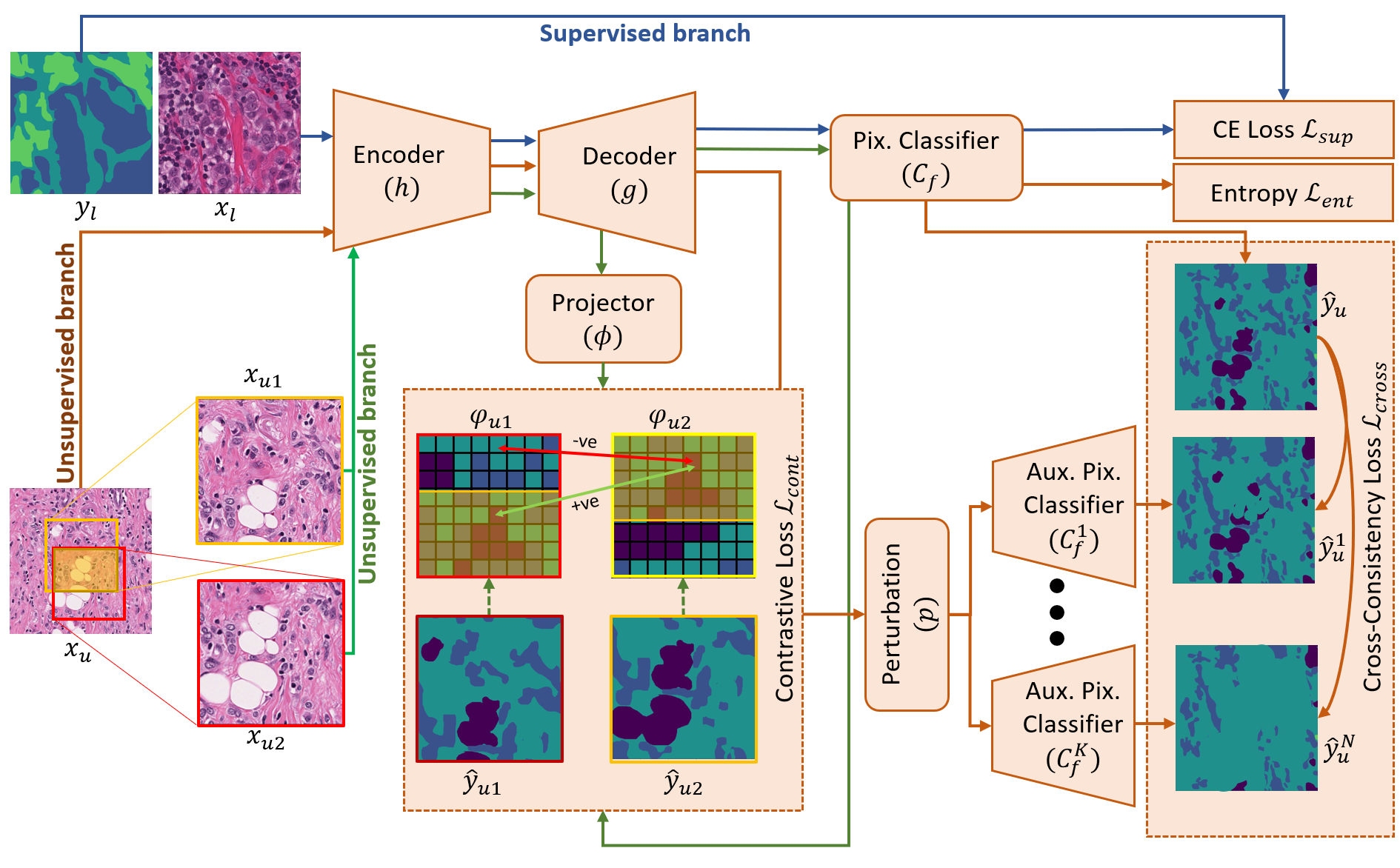}
    \caption{Overview of the proposed framework (CRCFP). The encoder and decoder are trained in a supervised manner with the cross-entropy (CE) loss for the labelled instances. For unlabelled instances, two cropped patches with partial overlap together with the input image were passed through the encoder, where the input image is used for contrastive and cross-consistency learning. }
    \label{fig:main_fig}
\end{figure}

\subsection{Semi-Supervised Semantic Segmentation}
SSL based semantic segmentation approaches utilise the aforementioned techniques to extract knowledge from unlabelled data. Recently, CutMix, MixUp, and CutOut based augmentation techniques were used togather with the student-teacher model where consistency was enforced between the mixed predictions \cite{french2019semi}. Guided collaborative training (GCT) by \cite{ke2020guided} performed network perturbations with the help of different network initialisation and enforced the dynamic consistency constraint between the predictions. Cross-consistency training (CCT) by \cite{ouali2020semi} performed perturbations on the main encoder's features and enforced consistency over the multiple decoders output making it robust to various perturbation types. Context-aware consistency by \cite{lai2021semi} proposed directional consistency loss for contrasting different contexts by cropping two overlapping patches of the same input to improve the representation learning. Recently, in the field of computational pathology, a few methods for semi-supervised semantic segmentation have been proposed. \cite{li2019signet} proposed a semi-supervised method for signet detection using with the help of self-supervised learning for label generation. \cite{lai2021semii} proposed a two stage SIM-FixMatch approach utilising self-supervised learning in the first stage and then using FixMatch for pseudo label generation along with consistency regularisation. \cite{cheng2020self} proposed an exponential moving average (EMA) student-teacher framework where the model is trained using the noisy labels to enforce the consistency over similar and dissimilar patch pairs. Cross-patch dense contrastive learning by \cite{wu2022cross} proposed a student-teacher based method to enforce EMA based consistency over predictions and to improve the internal representations. Pixel-wise contrastive loss was applied to background and foreground patches for improving the internal feature representations.

In this work, we show that (a) by enforcing consistency over varying contexts and feature perturbations in encoder's latent space, models can generalise better and (b) minimising entropy in output prediction maps can boost the confidence of the final predictions resulting in improved performance.

\section{The Proposed Method}

Figure \ref{fig:main_fig} shows an overview of the proposed framework (CRCFP), where $L = \{ (x_l^1,y_l^1),...,(x_l^n,y_l^n) : n \in [1, ..., N]\}$ represents the $N$ labelled images while $ U = \{ (x_u^1),...,(x_u^m) : m \in [1, ..., M]\} $ represents the $M$ unlabelled images. Labelled and unlabelled images $x_l$ and $x_u$ were sampled from $L$ and $U$ respectively in batches. Both images $x_l,x_u$ are of $H \times W \times D$ spatial dimensions with corresponding pixel-wise mask ${y_l} = {\mathbb{R}^{C \times H \times W}}$ only for labelled image where $C$ is the number of classes. Each labelled image $x_l$ is passed through the supervised pathway of the CRCFP framework (blue arrows in Figure \ref{fig:main_fig}) whereas the unlabelled images $x_u$ pass through the unsupervised pathways of the framework (brown arrows in Figure \ref{fig:main_fig}) along with two overlapping patches extracted randomly from $x_u$ denoted as $x_{u1},x_{u2}$ (green arrows in Figure \ref{fig:main_fig}). Feature maps are extracted from the input image using the shared encoder $h(\cdot; \theta_h)$ and decoder $g(\cdot; \theta_g)$ as $f(\cdot ; \theta_f)=h(\cdot ; \theta_h) \circ g(\cdot ; \theta_g)$ resulting in feature maps for each input as $f_l = f(x_l; \theta_f)$, $f_u = f(x_u; \theta_f)$, $f_{u1} = f(x_{u1}; \theta_f)$ and $f_{u2} = f(x_{u2}; \theta_f)$. Further, $f_l$ and $f_u$ are processed by a pixel classifier $C_f$ for final prediction as ${\hat y_l = C_f(f_l; \theta_p)}$ and ${\hat y_u = C_f(f_u; \theta_p)}$ where ${\hat y_l}$ is optimised using the cross-entropy loss over $y_l$ as $\mathcal{L}_{sup}$ shown in equation \ref{eq:loss_sup}.

\begin{equation}
    \label{eq:loss_sup}
    \mathcal{L}_{sup} =  - \frac{1}{N}\sum\limits_{c=1}^C{\sum\limits_{i=1}^N {{y_{i,c}}\log (\hat y_{i,c})}} 
\end{equation}

\begin{figure}
    \centering
    \includegraphics[width=0.9\textwidth]{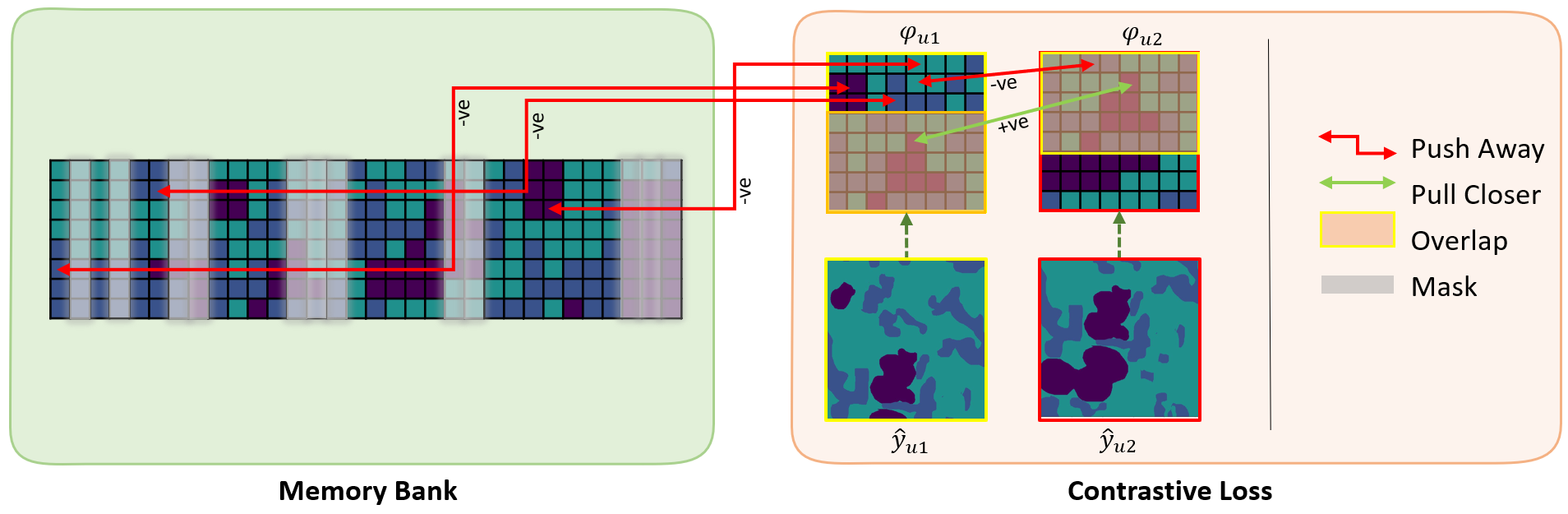}
    \caption{Directional contrastive loss working for context-aware consistency, where from $\varphi_{u1},\varphi_{u2}$ overlapping area (yellow overlay) positive pixels with higher confidence pull each other closer (orange arrows) while negative pixels from $\varphi_{u2}$ as well as from memory bank push each other apart (red arrows). Where class masks $\hat y_{u1},\hat y_{u2}$ (dashed green arrows) were applied to get the negative samples from $\varphi_{u2}$ and from the memory bank illustrated in the grey overlay. }
    \label{fig:context_aware_contrast}
\end{figure}

\subsection{Context-Aware Consistency}
With only the supervised loss $\mathcal{L}_{sup}$, the model may start relying excessively on contexts due to limited labelled data. Context-aware consistency can alleviate this issue by aligning the two different contexts of the same patch with the help of contrastive learning. For this purpose, encoded feature maps $f_{u1}$ and $f_{u2}$ are projected to a low-dimensional space using a non-linear projector $\phi$ to preserve important contextual information. The choice of non-linear projection head as compared to linear and identity projection head is due to its superior performance \cite{chen2020simple}. The projection head $\phi(\cdot;\theta_z)$ outputs projection maps as $\varphi_{u1} = \phi(f_{u1};\theta_z)$ and $\varphi_{u2} = \phi(f_{u2};\theta_z)$. Similar to \cite{lai2021semi}, context-aware consistency is maintained between the overlapping regions of $\varphi_{u1}$ and  $\varphi_{u2}$ using the directional contrastive loss $\mathcal{L}_{cont}$ to keep the feature representation consistent under different contexts as shown in \ref{fig:context_aware_contrast}. For computing directional consistency loss, first class maps $\hat y_{ui}$ were extracted using pixel classifier $C_f$ and then maximum probability among all classes $C$ is maintained using max probability as it is linked with higher confidence as shown in equation \ref{eq:y_hat_ui}.
\begin{equation}
    \label{eq:y_hat_ui}
    {\hat y_{ui}} = \mathop {\arg \max }\limits_{x{\text{ }} \in {\text{ }}U} {C_f}({f_{ui}}; \theta_p)
\end{equation}
where $i \in \{1,2\}$ and higher probability features are used to align less confident features towards the more confident features \cite{ke2020guided,lai2021semi,wu2022cross} which can improve learning by avoiding the exchange of unreliable knowledge from the less confident feature as shown in Figure \ref{fig:context_aware_contrast}. In order to extract negative samples (i.e., negative pairs), class maps as ${\hat y_{u1} = C_f(f_{u1}; \theta_p)}$ and ${\hat y_{u2} = C_f(f_{u2}; \theta_p)}$ were used. A positive feature projection $\varphi_{u1+}$ with class map $\hat y_{u1+}$ (i.e., in case of $u_1$ $\rightarrow$ $u_2$), the negative samples $\eta$ should have $(\hat y_{u1+} \ne \hat y_{u-})$ as shown in \ref{eq:neg_sample_map}. Further, to avoid less confident features from contributing towards the loss, a threshold $ \lambda $ is applied to avoid an exchange of knowledge between less confident features. The $\ell_{cont(\varphi_{u1},\varphi_{u2})}$ loss for one pair is calculated as shown below,

\begin{equation}
    \label{eq:loss_cont_1}
    \ell_{cont(\varphi_{u1},\varphi_{u2})} =
    - \frac{1}{M}\sum\limits_{h,w} {\mathcal{M}_{+}.\log \frac{{sim(\varphi_{u1},\varphi_{u2})}}{{sim(\varphi_{u1},\varphi_{u2}) + \sum\limits_{\varphi_{u-}  \in \eta } {\mathcal{M}_{u-} \cdot sim(\varphi_{u1},\varphi_{u-})} }}} 
\end{equation}

\begin{equation}
    sim({\varphi _{u1}},{\varphi _{u2}}) = \exp (\frac{{\varphi _{u1}^{\rm T}{\varphi _{u2}}}}{{\left\| {{\varphi _{u2}}} \right\|\left\| {{\varphi _{u2}}} \right\|\tau }})
\end{equation}

\begin{equation}
    \label{eq:neg_sample_map}
    {\mathcal{M}_ - } = \left\{ {\begin{array}{*{20}{l}}
      {1\;}&{{\text{ if }}\;{{\hat y}_{u1 + }} \ne {{\hat y}_{u- }} ,\;} \\ 
      {0\;}&{{\text{ otherwise}}{\text{.}}} 
    \end{array}} \right.
\end{equation}

\begin{equation}
{\mathcal{M}_ + } = \left\{ {\begin{array}{*{20}{l}}
  {\mathcal{M}_{c + }^{}}&{{\text{ if }}\;\max {C_f}({f_{u1}};{\theta _p}) > \lambda ,\;} \\ 
  {0\;}&{{\text{ otherwise}}{\text{.}}} 
\end{array}} \right.
\end{equation}

\begin{equation}
   {\mathcal{M}_{c + }} = \left\{ {\begin{array}{*{20}{l}}
  1&{{\text{ }}\max {C_f}({f_{u1}};{\theta _p}) < \max {C_f}({f_{u2}};{\theta _p}),\;} \\ 
  {0\;}&{{\text{ otherwise}}{\text{.}}} 
\end{array}} \right.
\end{equation}
where $sim(.)$ is the cosine similarity measure with temperature $\tau$, $\mathcal{M}_{c+}$ represents the binary mask for confident features corresponding to $\varphi_{u1+}$. $\mathcal{M}_{+}$ is the binary mask for positive confident samples above threshold $\lambda$. $\mathcal{M}_{-}$ is the binary mask for negative samples indicating different pseudo labels between $\varphi_{u+}$ and $\varphi_{u-}$. To increase the negative samples, we have used the memory bank which stores features from recent batches to further increase the negative samples for better contrastive performance \cite{chen2020simple,lai2021semi,wu2022cross}. Finally, the directional contrastive loss $\mathcal{L}_{cont}$ is calculated as below:

\begin{equation}
    \label{eq:loss_cont}
    \mathcal{L}_{cont} = \ell_{cont(\varphi_{u1},\varphi_{u2})} + \ell_{cont(\varphi_{u2},\varphi_{u1})}
\end{equation}

\subsection{Cross-Consistency Training}
 As context-aware consistency improves the model's robustness towards changing contexts without losing self-awareness, the model is still susceptible to small perturbations in the input due to limited labelled data. Therefore, in order to leverage unlabelled data and make the model invariant to small perturbations, we utilise the cross-consistency training \cite{ouali2020semi} where $f_u$ is perturbed $K$ times for each perturbation type and consistency is maintained between the output of pixel classifier and auxiliary classifiers. This not only improves the model's robustness but also regularises the main pixel classifier towards correct predictions. We use ${\hat y_u}$ to regularise the pixel classifier over the mean square error (MSE) loss by measuring the distance between the output of the main pixel classifier $C_f$ and the output of auxiliary classifiers $C_f^k$. Formally, a perturbation function $p_k$ with $k \in \{1,K\}$ perturbations outputs a perturbed version of the $f_u$ as $\hat f_u^k = p_k(\hat f_u)$ for a perturbation type and the cross-consistency training loss $\mathcal{L}_{cross}$ can be defined as below,

\begin{equation}
    \label{eq:loss_cross}
   {\mathcal{L}_{cross}} = \frac{1}{M}\frac{1}{K}\sum\limits_{{x_u} \in U} {\sum\limits_{k = 1}^K {d({{\hat y}_u},C_f^k(\hat f_u^k))} }
\end{equation}
where $d$ measures the squared distance between the output probabilities of the main pixel classifier and perturbed pixel classifier output. Following perturbations are applied to enforce the consistency:

\textbf{Feature Noise}: A uniformly sampled noise from the interval $[\alpha, \beta]$ is added to the features map $f_u$ in two steps. First sampled noise is multiplied with $f_u$ to scale the noise relative to feature activations. Second, the scaled sample noise is then added to the feature map $f_u$. This makes the noise proportional to each feature activation as shown below.

\begin{align}
    \Omega &\sim \mathcal{U}(\alpha , \beta )  \\
    f_u^{noise} &= ({f_u} \odot \Omega) + {f_u} 
\end{align}
    
\textbf{Feature Dropout}: A uniform sample threshold $\gamma$ is used to prune the less confident activations to stop the model from relying on those activations. This is done by first summing the $f_u$ over different channels and then normalising it using min-max normalisation resulting in $ f'_u$. Anything below $\gamma$ is dropped as seen below:

\begin{align}
    \gamma &\sim \mathcal{U}(\alpha ,\beta )  \\
       {\mathcal{M}_{drop}} &= \left\{ {\begin{array}{*{20}{l}}
      {1\;}&{{\text{ if }}\;{{f'}_u} < \gamma ,\;} \\ 
      {0\;}&{{\text{ otherwise}}{\text{.}}} 
    \end{array}} \right. \\
    f_u^{drop} &= {\mathcal{M}_{drop}} \odot {f'_u}
\end{align}
where $\mathcal{M}_{drop}$ is the binary mask containing threshold values for pruning the activations.

\textbf{DropOut}: A fraction of activations are dropped out spatially where the fraction is decided using the Bernoulli distribution with probability $\delta$. 

\begin{align}
    {\mathcal{M}_{dropout}} &\sim Bernoulli(\delta) \\
    f_u^{dropout} &= {\mathcal{M}_{dropout}} \odot {f_u}
\end{align}

\subsection{Entropy Minimisation} Context-aware contrastive learning and cross-consistency training improves the encoder's features but it often fails to improve the final pixel classifier leading to less reliable pseudo labels corrupting the training from unlabelled data. As higher confidence means better prediction maps resulting in more refined pseudo labels which can help train both context-ware and cross-training with improved positive/negative pairs and pseudo labels. Hence, in order to improve the confidence of predictions, we employ entropy regularisation following its applications in semi-supervised learning \cite{grandvalet2004semi,chen2019domain,liu2021domain,wu2022cross} as shown in \ref{eq:loss_ent} where it penalises the uncertain prediction in the unlabelled data and improves the overall confidence of the prediction maps.

\begin{equation}
    \label{eq:loss_ent}
    {\mathcal{L}_{ent}} =  - \frac{1}{M}\sum\limits_{m = 1}^M {\sum\limits_{c = 1}^C {{{\hat y}_u}} } \log {\hat y_u}  
\end{equation}


\subsection{Training}
Finally, the entire framework is trained in an end-to-end fashion using a weighted combination of the above mentioned losses as shown below,

\begin{equation}
    \label{eq:loss}
    \mathcal{L} = {w_{sup}}{\mathcal{L}_{sup}} + {w_{cont}}{\mathcal{L}_{cont}} + {w_{cross}}{\mathcal{L}_{cross}} + {w_{ent}}{\mathcal{L}_{ent}}
\end{equation}
where $w_{sup}$, $w_{cont}$, $w_{cross}$ and $w_{ent}$ correspond to the weights for each loss component respectively.

\section{Experiments}

\subsection{Datasets}

We evaluated the proposed framework on two publicly available datasets, the Breast Cancer Semantic Segmentation (BCSS) \cite{amgad2019structured} and Multi-organ Nucleus Segmentation Challenge (MoNuSeg) \cite{kumar2019multi} dataset for semantic segmentation. The data was obtained from the respective challenge pages hosted on Grand Challenge for Medical Image Analysis website (https://grand-challenge.org/).

\textbf{MoNuSeg}. The MoNuSeg challenge was organised as a MICCAI 2018 satellite event and contains 21,623 annotated nuclei from 30 H\&E stained images for training and contains 7223 annotated nuclei from 14 H\&E stained images for testing purposes. Annotations were done by engineering students and then an expert pathologist served as quality control for the annotated nuclei. Each image is of size 1000 $\times$ 1000 extracted from a WSI scanned at 40$\times$ resolution of an individual patient obtained from The Cancer Genome Atlas Program (TCGA) \cite{weinstein2013cancer}. WSIs are sampled from 18 different centres and 7 different organs including breast, liver, kidney, prostate, bladder, colon and stomach with various tumour stages. 

\textbf{BCSS}. The BCSS challenge was conducted in 2021 and contains over 20,000 annotated regions of interest (ROI) from 151 H\&E stained WSIs with the same number of patients from TCGA \cite{weinstein2013cancer}. 25 annotators including pathologists, residents, and medical students helped annotate this large scale data into 25 refined categories which are later merged into 5 broad categories as tumour, stroma, inflammatory, necrosis, and others. For this work, we have used the same 5 broad categories by relabelling the regions and then split them into training and test centres according to the \cite{amgad2019structured} where there were 14 centres for training and 7 centres for testing.

\subsection{Implementation Details}

\subsection{Data Preparation}
In order to validate the CRCFP framework, we evaluated it against different label proportions of each dataset. Where for BCSS different label proportions were collected from different centres (hospitals) to make training more susceptible to variation in colours enabling domain shift. DL methods often fail to perform well on samples from a different domain (centres), mainly due to domain shift, this also makes it a domain generalisation problem. Therefore, the training set was divided into portions by diving the total training centres as 1/1 (full), 1/2 (half), 1/4 (quarter), and 1/8 (one-eighth) centres where 1/8 results in training images coming from only 1 centre, while the test set remains intact as it is. Similarly, for 1/4 (quarter) training images comes from 4 centres and 7 centres for 1/2 (half). For MoNuSeg, different label proportions were based on training images themselves and are then divided into 1/1, 1/8, 1/16, and 1/32 proportions to make it comparable to the work of \cite{wu2022cross}. Further, this whole process is repeated using 3 different random seeds and then the results are reported using mean aggregation with standard deviation.

\subsection{Evaluation metrics}
In order to compare our proposed method quantitatively with other state-of-the-art methods (SOTA), we have used different quantitative measures including accuracy, F1-score (Dice) and mean intersection over union (mIoU) for both the datasets.

\subsection{Network Architecture}
We used DeepLab-v3 \cite{Chen2017DeepLab} as base segmentation network with ResNet-50 \cite{He2016Deep} encoder pretrained on ImageNet \cite{deng2009imagenet}. Where the projector consists of two fully connected (FC) layers of size 128 with ReLU as an intermediate activation layer, FC $\rightarrow$ ReLU $\rightarrow$ FC. Pixel classifiers consist of convolutional layers with a kernel of size 1 $\times$ 1 to reduce the number of channels to total classes with non-linear ReLU activation. The final layers upsamples the output using bi-linear interpolation to match the input size as $H \times W \times C$. 

\subsection{Experimental Settings}
The input size for the proposed framework for both labelled and unlabelled images was 320 $\times$ 320. For contrastive learning, two patches $x_{u1}$ and $x_{u2}$ were randomly cropped from the unlabelled image with an overlap in the range of [0.1, 1.0] and are resized to match the input dimensions. For positive filtering mask $\lambda$ was set to 0.75 and $\tau = 0.1$ as temperature for cosine similarity. For cross-consistency training, number of auxiliary pixel classifiers were set to $K = 4$ for each perturbation type and for feature noise perturbation the parameters $\alpha = -0.3$, $\beta = 0.3$ were used. For feature dropout perturbation, $\alpha = 0.75$, $\beta = 0.9$ were used as they can help remove approximately 10\% to 30\% of active regions from the feature map. Also, for simple Dropout the probability for Bernoulli distribution was set to  $\delta=$ 0.5. During training, a set of standard augmentations were applied to the input images including horizontal and vertical flipping, gaussian blur, colour and grey scaling. PyTorch was used for implementing this framework where for optimisation we train the whole framework for 80 epochs. For the initial 5 epochs, only supervised loss $\mathcal{L}_{sup}$ was used to train the whole framework as this provides a stable head start for the semi-supervised learning. The batch size of 8 was used for labelled and unlabelled images with stochastic gradient descent (SGD) optimiser and a learning rate of 0.001. As a common practice, \textit{poly} learning rate decay policy was used where the learning rate is scaled using $1 - {(\frac{{iter}}{{\max \_iter}})^{power}}$ at each iteration with power = 0.9. Weights with respect to different losses  $\mathcal{L}_{sup}$, $\mathcal{L}_{cont}$, $\mathcal{L}_{cross}$ and $\mathcal{L}_{end}$ were set to fixed values as $w_{sup} = 1$, $w_{cont} = 0.1$, $w_{cross} = 0.01$ and $w_{ent} = 0.01$ respectively. All models were trained with the same configurations for both datasets where two Nvidia GeForce 1080Ti GPUs are used for training.

\section{Results}
The performance of our proposed method (CRCFP) compared to recent SOTA semi-supervised semantic segmentation methods including DeepLab \cite{Chen2017DeepLab}, CCT \cite{ouali2020semi}, CAC \cite{lai2021semi} and CDCL \cite{wu2022cross} \footnote{CDCL \cite{wu2022cross} cannot be applied to the multi-class problem as it divides the patches into foreground and background only for contrastive learning.} is shown in Table \ref{tab:results_table_bcss} and Table \ref{tab:results_table_monuseg}. As these methods are implemented using different configurations and baseline segmentation models. For a fair comparison, we have implemented these methods within a unified framework with the same segmentation baseline, experimental settings and data augmentations.

\begin{table}
  \caption{Comparison of the state-of-the-art methods with mIoU, dice score and accuracy aggregated for 3 different random seeds as mean (standard deviation). The first column represents the fraction of data used for training the model. }
  \label{tab:results_table_bcss}
  \centering
  \begin{tabular}{lllll}
    \hline
    \multicolumn{5}{c}{BCSS}  \\
    \hline
    Fraction & Method & mIoU & Dice & Accuracy \\
    \hline
    1/8 & DeepLab-v3 \cite{Chen2017DeepLab}  & 40.99 (7.96)  & 55.96 (9.1)  & 66.53 (6.07)   \\
    1/8 & CCT \cite{ouali2020semi} & 22.84 (0.54)  & 32.01 (0.69)  & 56.14 (0.70)    \\
    1/8 & CAC \cite{lai2021semi} & 44.67 (6.32)  & 58.97 (7.51)  & 72.43 (3.40)   \\
    1/8 & CDCL \cite{wu2022cross}  & -  & -  & -  \\
    1/8 & CRCFP  &\textbf{ 47.09 (6.18)} &\textbf{ 61.84 (6.59)} & \textbf{73.20 (3.31)}  \\
    \hline
    1/4 & DeepLab-v3 \cite{Chen2017DeepLab}  & 53.03 (0.88)  & 68.52 (0.94)  & 75.70 (0.65)    \\
    1/4 & CCT \cite{ouali2020semi}  & 30.63 (2.19)  & 43.24 (3.33)  & 62.43 (0.89)    \\
    1/4 & CAC \cite{lai2021semi}  & 58.65 (0.65)  & 73.33 (0.55)  & 78.60 (0.42)   \\
    1/4 & CDCL \cite{wu2022cross}  & - & -  & -  \\
    1/4 & CRCFP  & \textbf{61.06 (0.98)} & \textbf{75.21 (0.74)} & \textbf{80.87 (0.89)} \\
    \hline
    1/2 & DeepLab-v3 \cite{Chen2017DeepLab}  & 56.26 (1.19)  & 71.33 (0.98)  & 78.07 (1.031) \\
    1/2 & CCT \cite{ouali2020semi}  & 29.78 (2.56)  & 41.63 (4.28)  & 61.94 (0.17)  \\
    1/2 & CAC \cite{lai2021semi}  & 60.44 (1.48)  & 74.67 (1.16)  & 80.50 (0.92) \\
    1/2 & CDCL \cite{wu2022cross}  & -  & - &- \\
    1/2 & CRCFP  & \textbf{61.86 (0.63)}  & \textbf{75.73 (0.59)}  & \textbf{81.18 (0.27)}  \\
    \hline
    1/1 & DeepLab-v3 \cite{Chen2017DeepLab}  & 61.29 (0.26)  & 75.49 (0.12)  & 81.10 (0.09) \\
    \hline
  \end{tabular}
\end{table}

\begin{table}
    \caption{Comparison of the state-of-the-art methods with mIoU, dice score and accuracy aggregated for 3 different random seeds as mean (standard deviation). The first column represents the fraction of data used for training the model.}
    \label{tab:results_table_monuseg}
    \centering
    \begin{tabular}{lllll}
    \hline
    \multicolumn{5}{c}{MoNuSeg}                \\
    \hline
    Fraction & Method & mIoU & Dice & Accuracy \\
    \hline
    1/32 & DeepLab-v3 \cite{Chen2017DeepLab}  & 60.09 (2.07)  & 73.89 (1.77)  & 79.45 (1.40)   \\
    1/32 & CCT \cite{ouali2020semi} & 41.13 (0.06)  & 50.31 (0.29)  & 74.33 (0.35)    \\
    1/32 & CAC \cite{lai2021semi} & 67.40 (1.12)  & 79.33 (0.90)  & 86.14 (0.62)   \\
    1/32 & CDCL \cite{wu2022cross} & 62.72  (1.83) & 75.95 (75.95)  & 81.66 (1.57)  \\
    1/32 & CRCFP  &\textbf{ 71.72 (0.22)} &\textbf{ 82.60 (0.24)} & \textbf{88.86 (0.23)}  \\
    \hline
    1/16 & DeepLab-v3 \cite{Chen2017DeepLab}  & 56.20 (5.76)  & 70.80 (4.58)  & 75.27 (75.27)    \\
    1/16 & CCT \cite{ouali2020semi}  & 40.99 (0.08)  & 49.56 (0.29)  & 75.17 (0.36)    \\
    1/16 & CAC \cite{lai2021semi}  & 71.44 (1.11)  & 82.47 (0.92)  & 88.27 (0.11)   \\
    1/16 & CDCL \cite{wu2022cross}  & 60.63 (1.15) & 74.40 (0.90)  & 79.99 (1.37)  \\
    1/16 & CRCFP  & \textbf{72.08 (2.07)} & \textbf{82.91 (1.52)} & \textbf{88.58 (1.16)} \\
    \hline
    1/8 & DeepLab-v3 \cite{Chen2017DeepLab}  & 59.67 (2.99)  & 73.59 (2.32)  & 78.98 (2.89) \\
    1/8 & CCT \cite{ouali2020semi}  & 40.9 (0.13)  & 50.00 (0.54)  & 74.63 (0.42)  \\
    1/8 & CAC \cite{lai2021semi}  & 74.56 (0.42)  & 84.73 (0.30)  & 89.91 (0.23) \\
    1/8 & CDCL \cite{wu2022cross}  & 57.07 (1.45)  & 71.63 (1.14) & 76.29 (1.48) \\
    1/8 & CRCFP  & \textbf{75.57 (0.85)}  & \textbf{85.19 (0.54)}  & \textbf{90.28 (0.42)}  \\
    \hline
    1/1 & DeepLab-v3 \cite{Chen2017DeepLab}  & 71.29 (0.16)  & 82.49 (0.11)  & 87.52 (0.09) \\
    \hline
\end{tabular}
\end{table}

\begin{figure}
    \centering
    \includegraphics[width=0.99\textwidth]{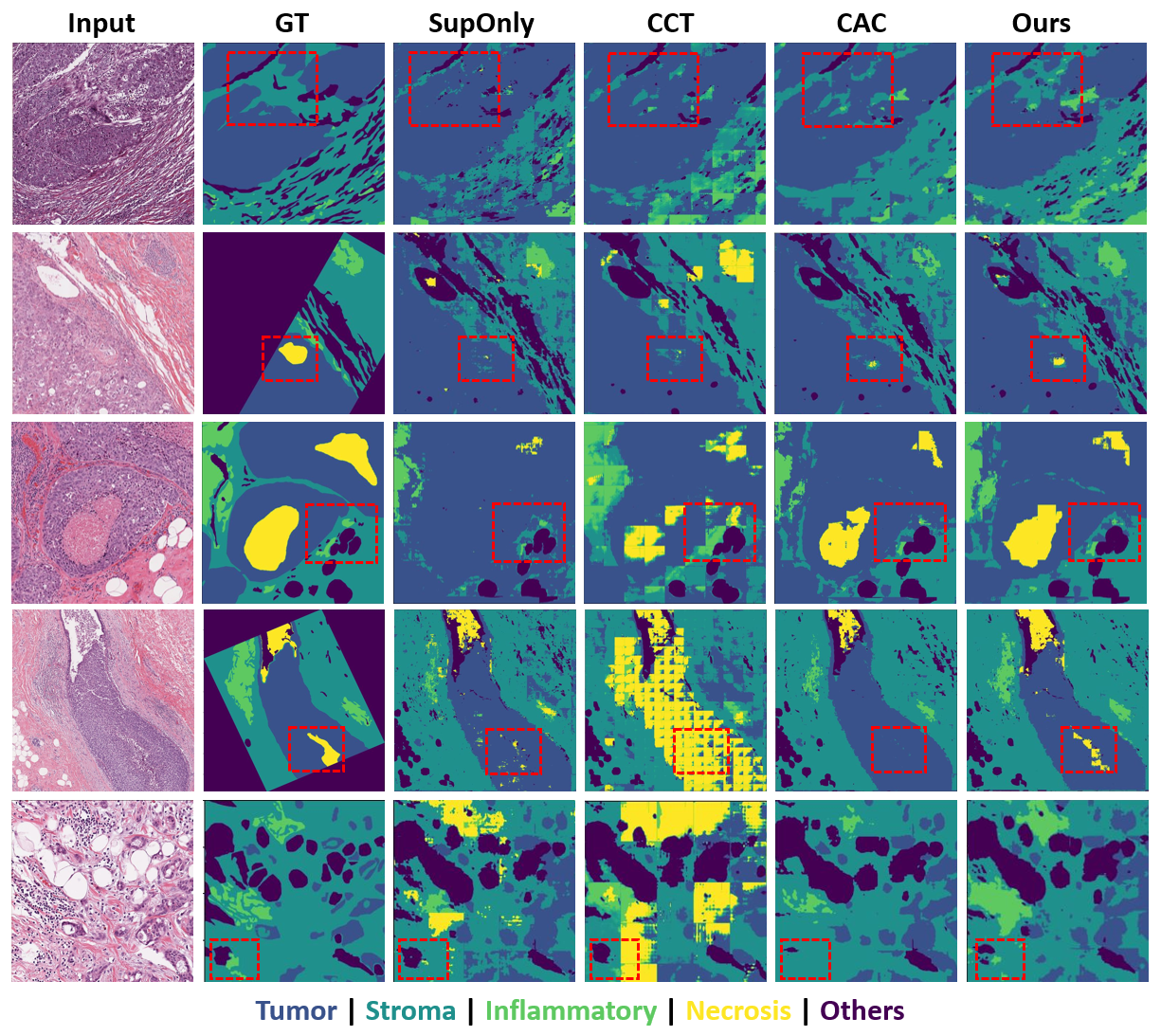}
    \captionsetup{width=0.99\textwidth}
    \caption{Visual comparison of the CRCFP with different state-of-the-art methods for tissue region segmentation with 1/2 training data only. Dashed red box highlights superior performance of our method as compared to SOTA methods.}
    \label{fig:bcss_pred}
\end{figure}

\begin{figure}
    \centering
    \includegraphics[width=0.99\textwidth]{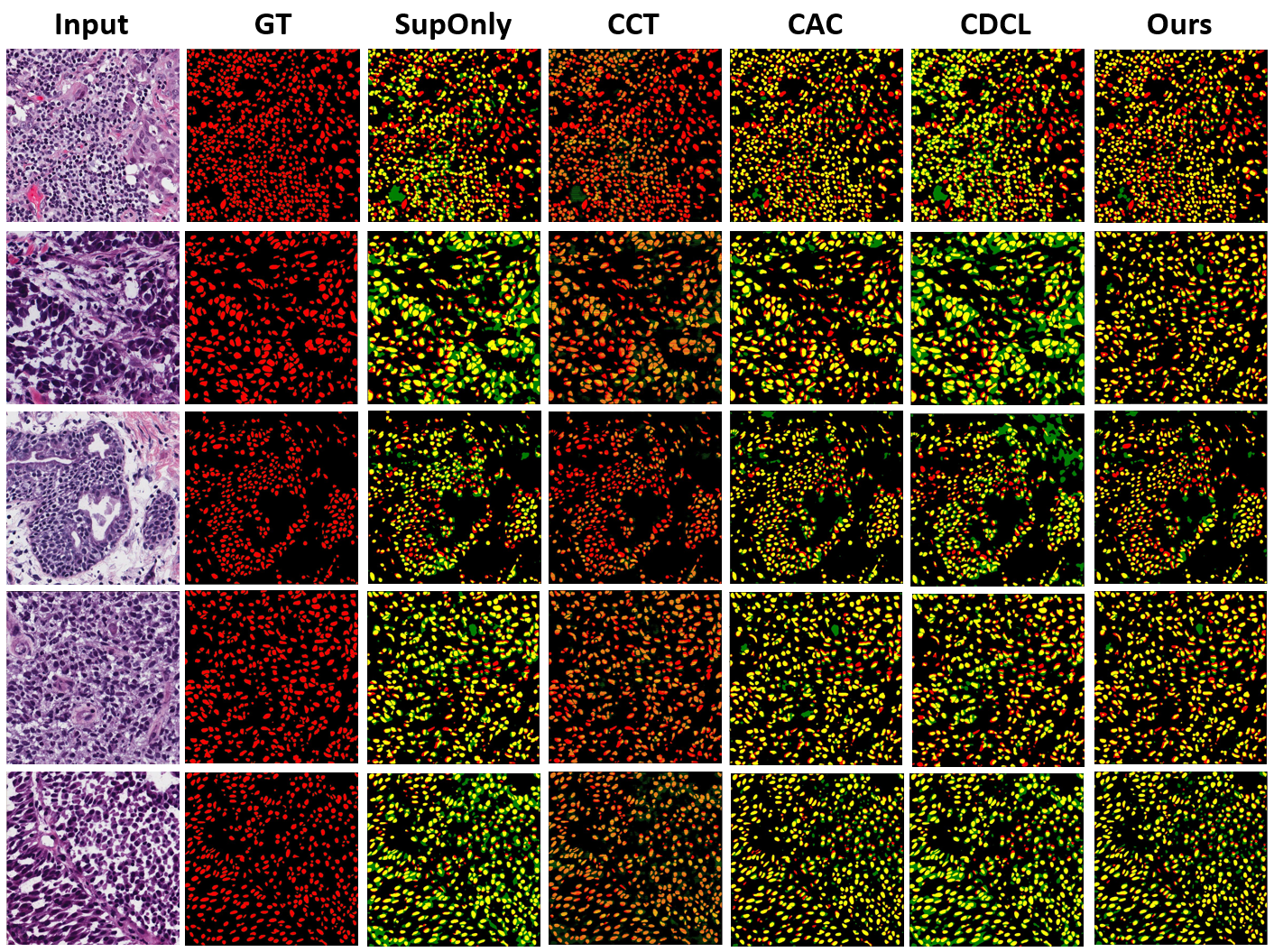}
    \captionsetup{width=0.99\textwidth}
    \caption{Visual comparison of the CRCFP with different state-of-the-art techniques in nuclei image segmentation with 1/8 training data only. GT represents the ground truth nuclei masks, and SupOnly shows the models trained with labelled training data only. Red pixels correspond to the ground truth while green shows the prediction. Yellow pixels represent the overlap regions between the prediction and ground truth.}
    \label{fig:monuseg_pred}
\end{figure}

Table \ref{tab:results_table_bcss} shows the performance of our CRCFP model compared to supervised and semi-supervised methods on all matrices for the BCSS dataset. Particularly, when 1/8 proportion of the training centres was used, it can be seen that in terms of mIoU our method performs $\sim$6\%  better than the supervised method and $\sim$3\% better than the recent CAC \cite{lai2021semi}. Similarly, its worth noting that with 1/4 of the total centres, the CRCFP performance is almost similar to fully supervised method with all data. On the other hand, the poor performance of CCT \cite{ouali2020semi} can be attributed towards heavy perturbations applied directly to the features where it brings perturbed features from different contexts closer without pushing dissimilar apart whereas CAC \cite{lai2021semi} not only bring them closes but also pushes away the features from different classes. However, it focuses more on encoder feature generalisation leaving pixel-classifier with less confident features.  Figure \ref{fig:bcss_pred} shows visual comparison of CRCFP with the SOTA algorithms, where, it can be observed that prediction maps of CRCFP are better as compared to the rest, specially highligted in the dashed red boxes.

Table \ref{tab:results_table_monuseg} shows the performance of CRCFP surpassing other SOTA methods in all data proportions and metrics, especially in 1/32 proportion of the MoNuSeg dataset. It can be seen that our CRCFP outperforms the CAC \cite{lai2021semi} by 4.32\% in mIoU with a smaller standard deviation of 0.22. It can also be observed that fully supervised models are more susceptible to domain generalisation problem from the table as in 1/32 proportion of the training images the performance of DeepLab-v3 \cite{Chen2017DeepLab} is 4\% better than the 1/16 proportion of the training images whereas there is more data available in the latter. This is due to the fact that in a random sampling of training images some training images are better indicators of the testing distribution due to similarities in the same stain, organ and tumour stage. However, most of the SOTA semi-supervised algorithms solve this issue with the help of unlabelled data as it can be seen that the performance increase with the increase in data for all these methods. Figure \ref{fig:monuseg_pred} shows a visual comparison of CRCFP with SOTA methods where it can be seen that our approach predicts fewer false positives as compared to CDCL \cite{wu2022cross}.

Further, in order to validate the contribution of each component (i.e., context-aware consistency, cross-consistency training and entropy minimisation) we conducted an extensive ablation study. The ablation study is performed on the BCSS dataset due to its complexity and multi-class nature, where we studied the effect of using all data proportions for the different encoders and in stripping the framework. While studying the effect of negative samples and the number of auxiliary pixel classifiers we used 1/8 data proportion.

\subsection{Encoder}
To verify the performance boost by plugging in a bigger encoder in the base segmentation network, we replaced ResNet-50 with ResNet-101 for all data proportions. Table \ref{tab:results_table_bcss101} shows the performance of the proposed CRCFP framework with a bigger encoder and it can be seen that there is a performance boost overall for most of the methods, especially for CCT \cite{ouali2020semi}. However, it can be observed that CRCFP with a smaller encoder (i.e., ResNet-50) still performs comparable/better than other SOTA techniques with a bigger encoder e.g., in 1/8 proportion CAC \cite{lai2021semi} with ResNet-101 achieves mIoU of 46.91 where CRCFP with ResNet-50 achieves mIoU of 47.09  showing superiority of our proposed method. Also, it is worth mentioning that with ResNet-101 the standard deviation we observed with ResNet-50 was reduced, owing to the fact that bigger encoders are more stable for semi-supervised learning frameworks. Overall the CRCFP framework provides improved and stable performance with bigger encoders as compared to the other methods.   

\begin{table}
    \captionsetup{width=.99\textwidth}
  \caption{Comparison of the state-of-the-art methods on mean (standard deviation) of mean intersection of union (mIoU), dice score and accuracy with baseline encoder as ResNet-101. The first column represents the fraction of data used for training the model.}
  \label{tab:results_table_bcss101}
  \centering
  \begin{tabular}{lllll}
    \hline
    \multicolumn{5}{c}{BCSS}  \\
    \hline
    Fraction & Method & mIoU & Dice & Accuracy \\
    \hline
    1/8 & DeepLab-v3 \cite{Chen2017DeepLab}  & 37.50  (6.61)  & 51.73  (7.51)  & 64.89  (5.92)   \\
    1/8 & CCT \cite{ouali2020semi} & 31.71  (4.64)  & 45.66  (5.96)  & 59.42  (3.46)    \\
    1/8 & CAC \cite{lai2021semi} & 46.91  (6.79)  & 61.92  (6.74)  & 72.01  (3.85)   \\
    1/8 & CRCFP  &\textbf{ 47.15  (6.76)} &\textbf{ 61.27  (7.72)} & \textbf{72.57  (2.82)}  \\
    \hline
    1/4 & DeepLab-v3 \cite{Chen2017DeepLab}  & 55.18  (1.88)  & 70.30  (1.70)  & 77.37  (1.25)    \\
    1/4 & CCT \cite{ouali2020semi}  & 42.63  (0.98)  & 58.35  (1.26)  & 66.94  (0.73)    \\
    1/4 & CAC \cite{lai2021semi}  & 61.48  (0.73)  & 75.52  (0.47)  & 80.78  (0.84)   \\
    1/4 & CRCFP  & \textbf{62.01  (0.40)} & \textbf{75.94  (0.29)} & \textbf{81.18  (0.49)} \\
    \hline
    1/2 & DeepLab-v3 \cite{Chen2017DeepLab}  & 60.37 (1.89)  & 74.5 (1.58)  & 80.57 (0.86) \\
    1/2 & CCT \cite{ouali2020semi}  & 44.01 (0.65)  & 59.64  (0.55)  & 67.66  (1.33)  \\
    1/2 & CAC \cite{lai2021semi}  & 61.95  (0.72)  & 75.77  (0.67)  & 81.09  (0.27) \\
    1/2 & CRCFP  & \textbf{63.01 (0.09)}  & \textbf{76.57 (0.09)}  & \textbf{81.67 (0.12)}  \\
    \hline
    1/1 & DeepLab-v3 \cite{Chen2017DeepLab}  & 62.33 (1.04)  & 76.22 (0.73)  & 81.68 (0.58) \\
    \hline
  \end{tabular}
\end{table}

\subsection{Network Schemes}
We validated the contribution of each component by breaking down the whole framework with respect to different losses and called them network schemes. We started with a baseline segmentation network i.e., DeepLab-v3 with ResNet-50 as SupOnly, Scheme.1 consists of using context-aware consistency loss, Scheme.2 consists of using context-aware consistency loss with entropy minimisation and finally Scheme.3 is our proposed framework with context-aware consistency loss with cross-consistency training and entropy minimisation. Table \ref{tab:results_table_bcss_breakdown} shows the schemes with respect to their respective losses being used, it can be seen that with each component's addition we can see improvement in overall performance. E.g., in 1/8 data proportion, the addition of context-aware consistency brings about 4\% of improvement while entropy minimisation further bumps it up by 1\% and finally cross-consistent training beings about 2\% of improvement accumulating the overall performance to $\sim$7\% from baseline supervised model. Also, for other data proportions the performance boost is not that much significant with the addition of these Scheme.2 and Scheme.3 as compared to Scheme.1. However, its worth mentioning that the standard deviation of Scheme.2 and Scheme.3 as compared to Scheme.1 is smaller which is due to the fact that these schemes brings confidence in prediction maps thus improving the overall performance with stability.

\begin{table}
  \caption{CRCFP breakdown in different Schemes with respect to their loss functions. SupOnly correspond to baseline segmentation model with $\mathcal{L}_{sup}$ loss only. Scheme.1 corresponds to addition of $\mathcal{L}_{cont}$ loss on top of SupOnly. Scheme.2 corresponds to addition of $\mathcal{L}_{ent}$ on top of Scheme.1 and finally Scheme.3 is addition of $\mathcal{L}_{cons}$ on top of Scheme.2.}
  \label{tab:results_table_bcss_breakdown}
  \centering
  \begin{tabular}{lllllll}
    \hline
    Method & Split & $\mathcal{L}_{sup}$ & $\mathcal{L}_{cont}$ & $\mathcal{L}_{ent}$ & $\mathcal{L}_{cons}$ & mIoU \\
    \hline
    SupOnly & 1/8 & \checkmark & $\times$ & $\times$ & $\times$ & 40.99 (7.96)    \\
    Scheme.1  & 1/8 & \checkmark  & \checkmark  & $\times$ & $\times$ & 44.67  (6.32)    \\
    Scheme.2 & 1/8 & \checkmark  & \checkmark  & \checkmark & $\times$ & 45.76  (6.12)   \\
    Scheme.3   & 1/8 & \checkmark  & \checkmark  & \checkmark & \checkmark & \textbf{47.09 (6.18)}  \\
    \hline
    SupOnly & 1/4 & \checkmark & $\times$ & $\times$ & $\times$ & 53.03 (0.88)   \\
    Scheme.1  & 1/4 & \checkmark  & \checkmark  & $\times$ & $\times$ & 58.65 (0.65)    \\
    Scheme.2 & 1/4 & \checkmark  & \checkmark  & \checkmark & $\times$ & 59.97  (1.47)   \\
    Scheme.3   & 1/4 & \checkmark  & \checkmark  & \checkmark & \checkmark & \textbf{61.06 (0.98)}  \\
    \hline
    SupOnly & 1/2 & \checkmark & $\times$ & $\times$ & $\times$ & 56.26 (1.19)    \\
    Scheme.1  & 1/2 & \checkmark  & \checkmark  & $\times$ & $\times$ & 60.44 (1.48)    \\
    Scheme.2 & 1/2 & \checkmark  & \checkmark  & \checkmark & $\times$ & 60.87  (1.39)  \\
    Scheme.3   & 1/2 & \checkmark  & \checkmark  & \checkmark & \checkmark & \textbf{61.86 (0.63)}  \\
    \hline
  \end{tabular}
\end{table}

\subsection{Negative Samples}
\begin{figure}
    \centering
    \includegraphics[width=0.75\textwidth]{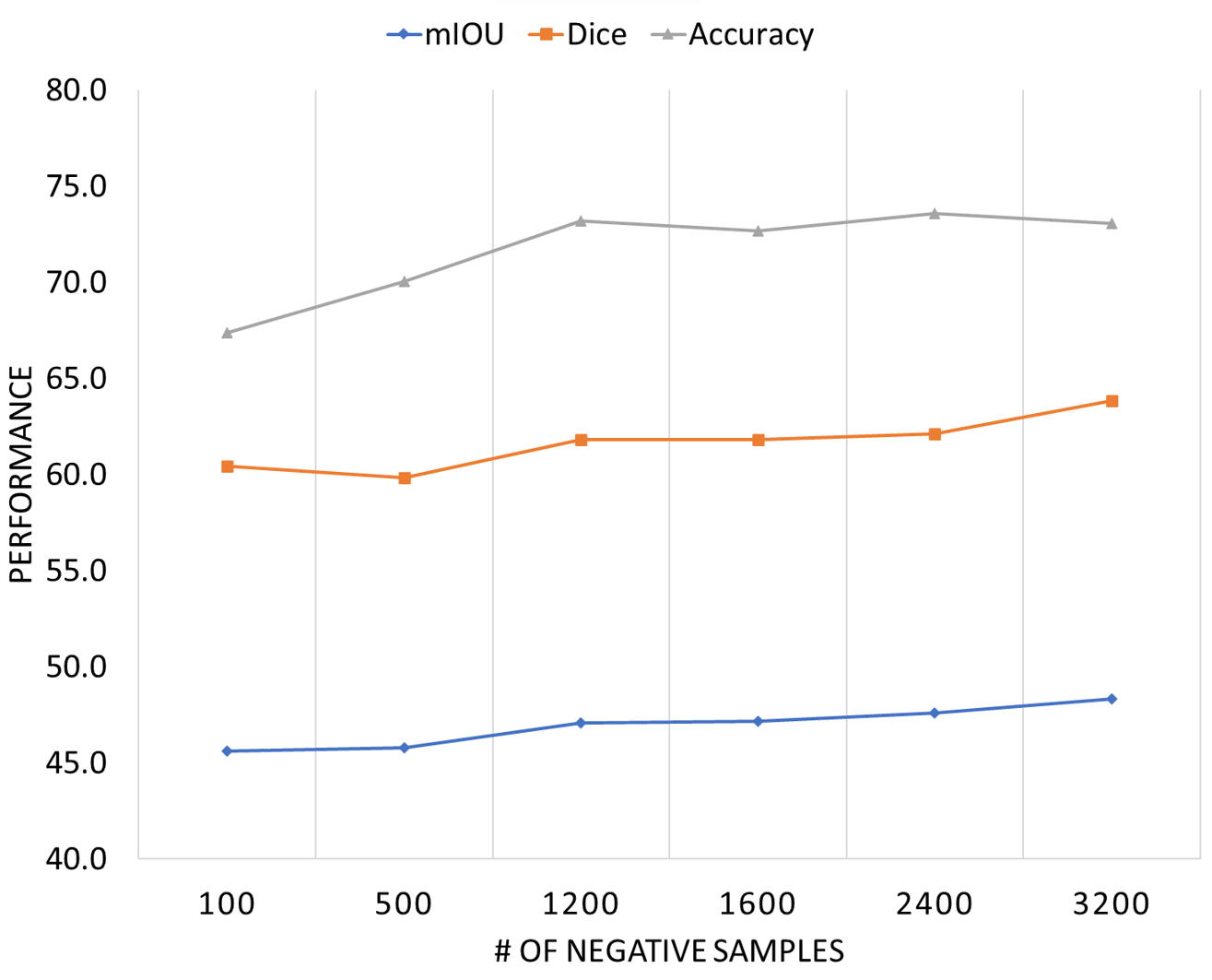}
    \caption{Performance graph with respect varying number of negatives samples used while training $\mathcal{L}_{cont}$ loss with BCSS data split of 1/8}
    \label{fig:bcss_neg_samples}
\end{figure}

As increasing the negative samples in training contrastive learning framework boosts the performance of the underlying model. This is done mostly by increasing the batch size to 2048 or 4096 where possible as the bigger the batch size the more samples you get for comparisons \cite{chen2020simple,caron2020unsupervised}. However, where it is not possible, another workaround is to use a memory bank where negative samples from previous batches were stored for later use. Therefore, in order to get the upper bound of performance in our framework with respect to negative samples, we have experimented with different number of negative samples as seen in Table \ref{tab:results_table_bcss_negative_samples}. It can be noticed that with increasing negative samples, the performance increases for a while and then it reaches the plateau and then increases with very little gain as it can also be observed visually in Figure \ref{fig:bcss_neg_samples}. This can be due to the fact that there might not be many variations to cover in the training set with more negative samples, thus reaching stable performance or very little performance gain. Also, due to gradient checkpoint functionality in PyTorch adding more negative samples does not effect the training efficiency drastically but does consume more compute time and memory. Hence, based on these observations, for this study, we set the number of negative samples to 1200 for its memory vs accuracy trade-off.

\begin{table}
  \caption{Performance of CRCFP with respect different number of negatives samples used while training $\mathcal{L}_{cont}$ loss with BCSS data split of 1/8}
  \label{tab:results_table_bcss_negative_samples}
  \centering
  \begin{tabular}{llll}
    \hline
    \#   & mIoU & Dice & Accuracy \\
    \hline
    100   & 45.62  (8.10) & 60.46  (8.87)  & 67.38  (8.14)   \\
    500  & 45.81  (7.78)  & 59.86  (8.88)  & 70.05  (5.30)    \\
    1200  & 47.09  (6.18)  & 61.84  (6.59)  & 73.20  (3.31)   \\
    1600  & 47.16  (6.70) & 61.81  (7.05) & 72.68  (3.07)  \\
    2400   & 47.60  (6.09) & 62.14  (6.80) & \textbf{73.58  (3.43)}  \\
    3200   & \textbf{48.34  (5.25)} & \textbf{63.83  (5.01)} & 73.06  (3.59)  \\
    \hline
  \end{tabular}
\end{table}

\subsection{Auxiliary Pixel Classifier}
\begin{figure}
    \centering
    \includegraphics[width=0.75\textwidth]{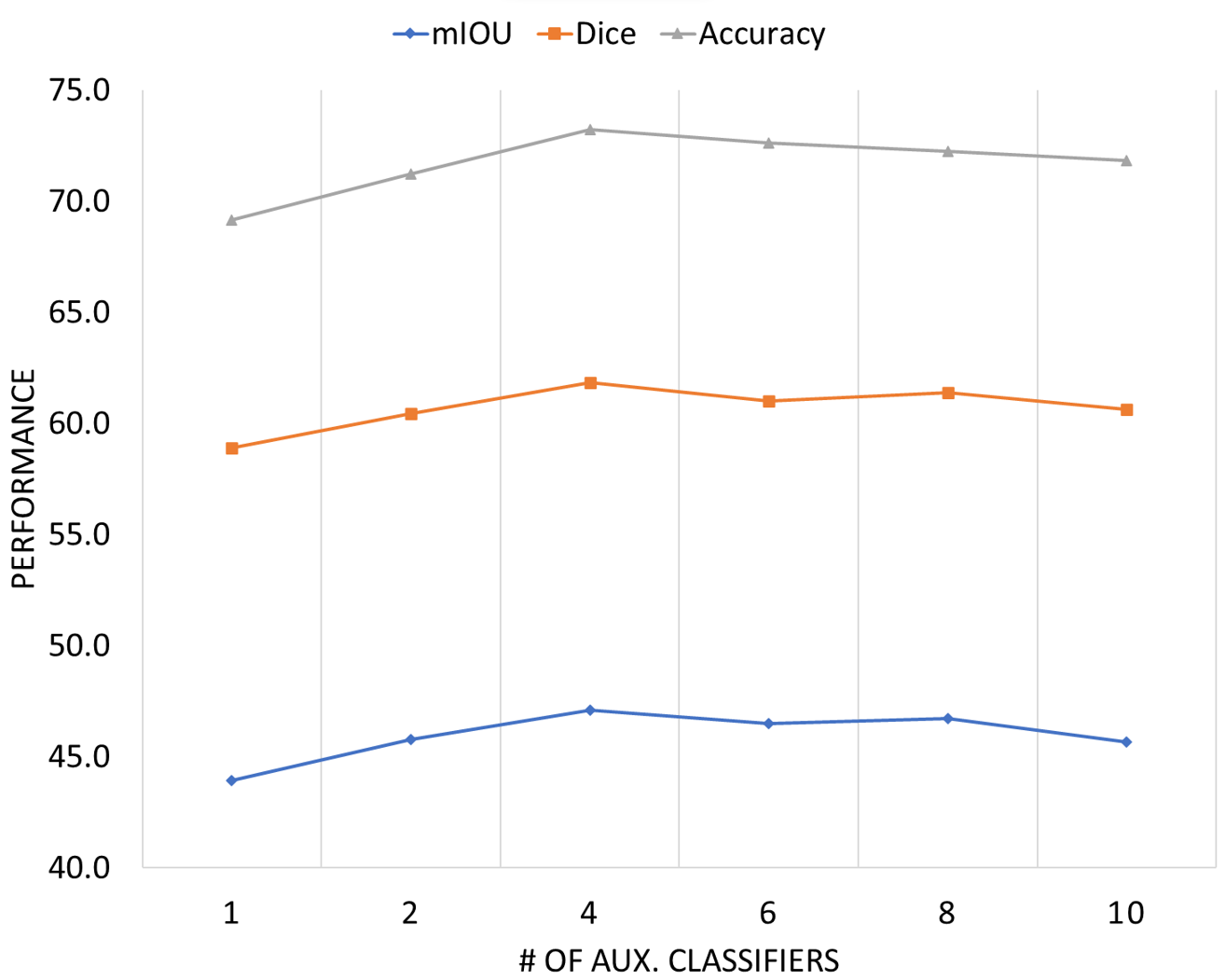}
    \caption{Performance graph with respect varying number of pixel classifiers used while training $\mathcal{L}_{cons}$ loss with BCSS data split of 1/8}
    \label{fig:bcss_aux_clfs}
\end{figure}

To see the effect of a varying number of auxiliary pixel classifiers with respect to different perturbations we conducted experiments with $K \in \{1,2,4,6,8,10\}$ as seen in Table \ref{tab:results_table_bcss_aux_classifiers}. It can be seen that increasing the number of pixel classifiers per perturbation increases the performance but the upper bound is achieved soon after it reaches $K = 4$, from where the performance drops slightly as can be observed in the Figure \ref{fig:bcss_aux_clfs}. Increasing the number of perturbations can result in more aggressive penalisation of the model overall as it accumulates to $K \times 3$ losses which can deviate the model from learning meaningful representations. Based on this observation we set the number $K = 4$ for our study for the rest of the comparisons for both datasets.

\begin{table}
  \caption{Performance of CRCFP with respect different number of $K$ auxiliary classifiers used while training $\mathcal{L}_{cons}$ loss with BCSS data split of 1/8}
  \label{tab:results_table_bcss_aux_classifiers}
  \centering
  \begin{tabular}{llll}
    \hline
    \#  & mIoU & Dice & Accuracy \\
    \hline
    1   & 43.94  (7.95) & 58.9  (8.27)  & 69.14  (5.54)   \\
    2  & 45.76  (7.51)  & 60.44  (7.88)  & 71.23  (4.23)    \\
    4  & \textbf{47.09  (6.18)}  & \textbf{61.84  (6.59)}  & \textbf{73.20  (3.31)}   \\
    6  & 46.48  (6.26) & 61.01 (6.73) & 72.60  (3.68)  \\
    8   & 46.72  (6.88) & 61.38  (7.29) & 72.25  (3.89)  \\
    10   & 45.68  (6.79) & 60.64  (7.20) & 71.84  (3.99)  \\
    \hline
  \end{tabular}
\end{table}

\section{Discussion}
Interpretable features from histology slides can be extracted by segmenting objects/structures from ROIs  e.g., nuclei, glands, stroma, tumours etc. Intrepretable features can enable discovery of novel digital bio-markers with explanations for histology images for hard tasks like survival analysi \cite{lu2018nuclear,shaban2022digital} and mutation prediction \cite{kather2020pan,diao2021human,bilal2021development}. Therefore, it is vital for the downstream tasks to have good quality and precise segmentation of region of interests. For this purpose, utilising unlabelled data for representation learning not only improves performance but also improves the internal representations for better learning. The qualitative and quantitative results along with the ablation study has shown superior performance of our proposed CRCFP with respect to other SOTA methods. However, it's worth exploring internal representations of the learned models (i.e.,feature embeddings) to account for (1) Consistency in feature space (2) Cluster assumption, for the sake of validation of aforementioned claims in the introduction section.

\begin{figure}[!ht]
    \centering
    \includegraphics[width=0.99\textwidth]{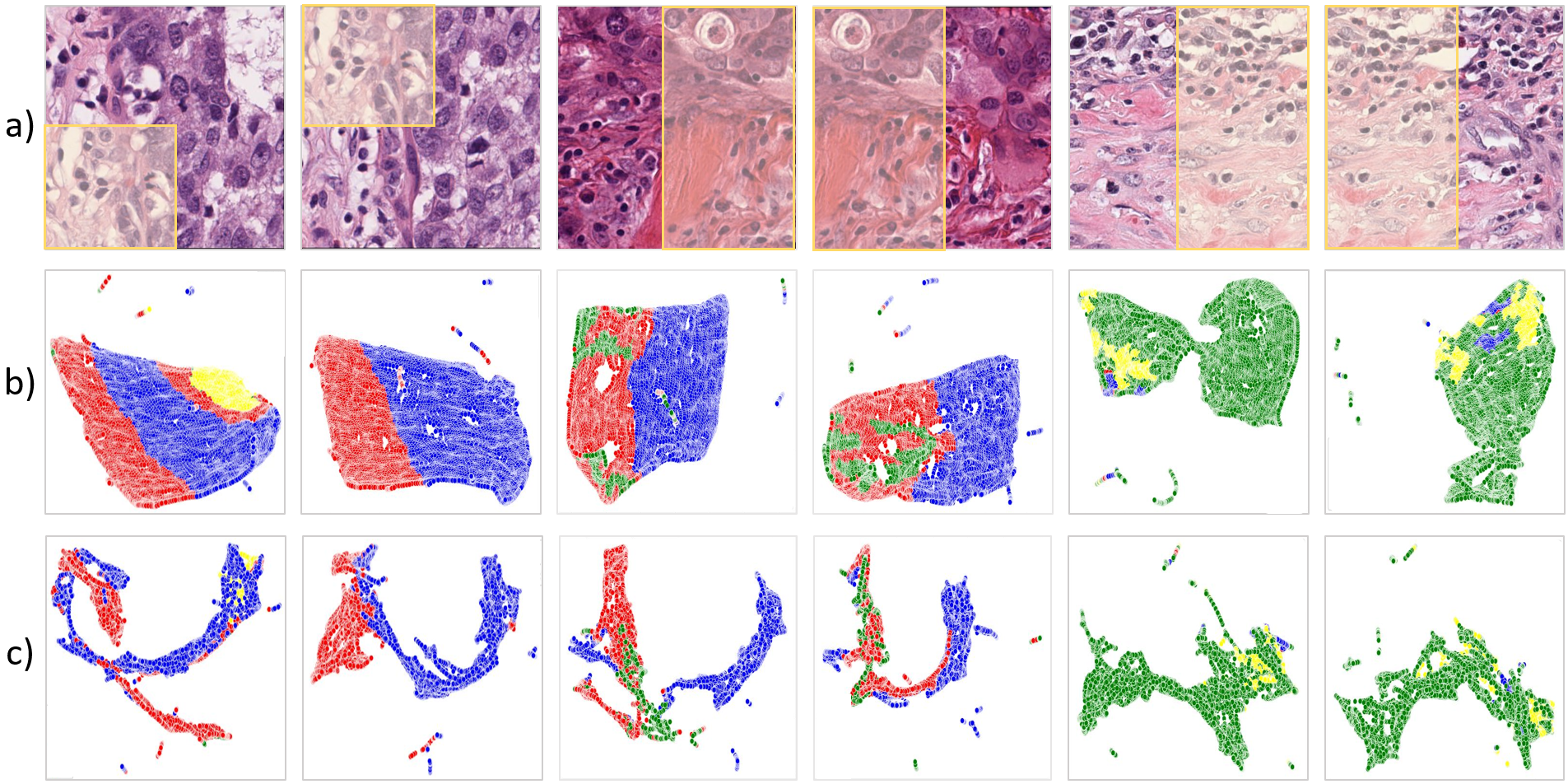}
    \captionsetup{width=0.99\textwidth}
    \caption{(a) BCSS dataset images with overlapping regions cropped sequentially from the same image to mimic changing contexts. (b) UMAP visualisations of features embedding distributions extracted from a fully supervised model. (c) UMAP visualisations of feature embedding distributions extracted from a semi-supervised model. Note that the feature embeddings are represented in the same UMAP space where dots with same colour represents feature embedding from the same class.}
    \label{fig:full_context_umap}
\end{figure}

\subsection{Feature Space Visualisation}
In order to observe the consistency in feature space, feature embeddings were extracted from both our SSL based CRCFP trained on 1/2 proportion of the training data vs DeepLab-v3 trained on all data (i.e., fully supervised), since they achieved same performance. Extracted feature maps were upsampled to match the size of the input image (i.e., 320 $\times$ 320) and are then mapped to lower dimensions using UMAP \cite{mcinnes2018umap} for visualisation purposes. It can be seen in Figure \ref{fig:full_context_umap} that the feature embedding distributions are consistent with varying contexts specially in the $1^{st}$ and $2^{nd}$ column for our CRCFP model as compared to fully supervised ones. Similarly, it can be observed in the other examples where the varying context is inherent due to the sequential overlap in patch tessellation process. Whereas, the fully supervised model is susceptible to perturbations in contextual cues as can be observed. It is worth noting the last two columns where the shape of feature embedding distribution changes along with the orientation of same samples points from the same class. Specially, the ones shown in yellow dots as compared to our proposed framework where the distributions are almost consistent under these perturbations.

\subsection{Cluster Assumption}
\begin{figure}
    \centering
    \includegraphics[width=0.99\textwidth]{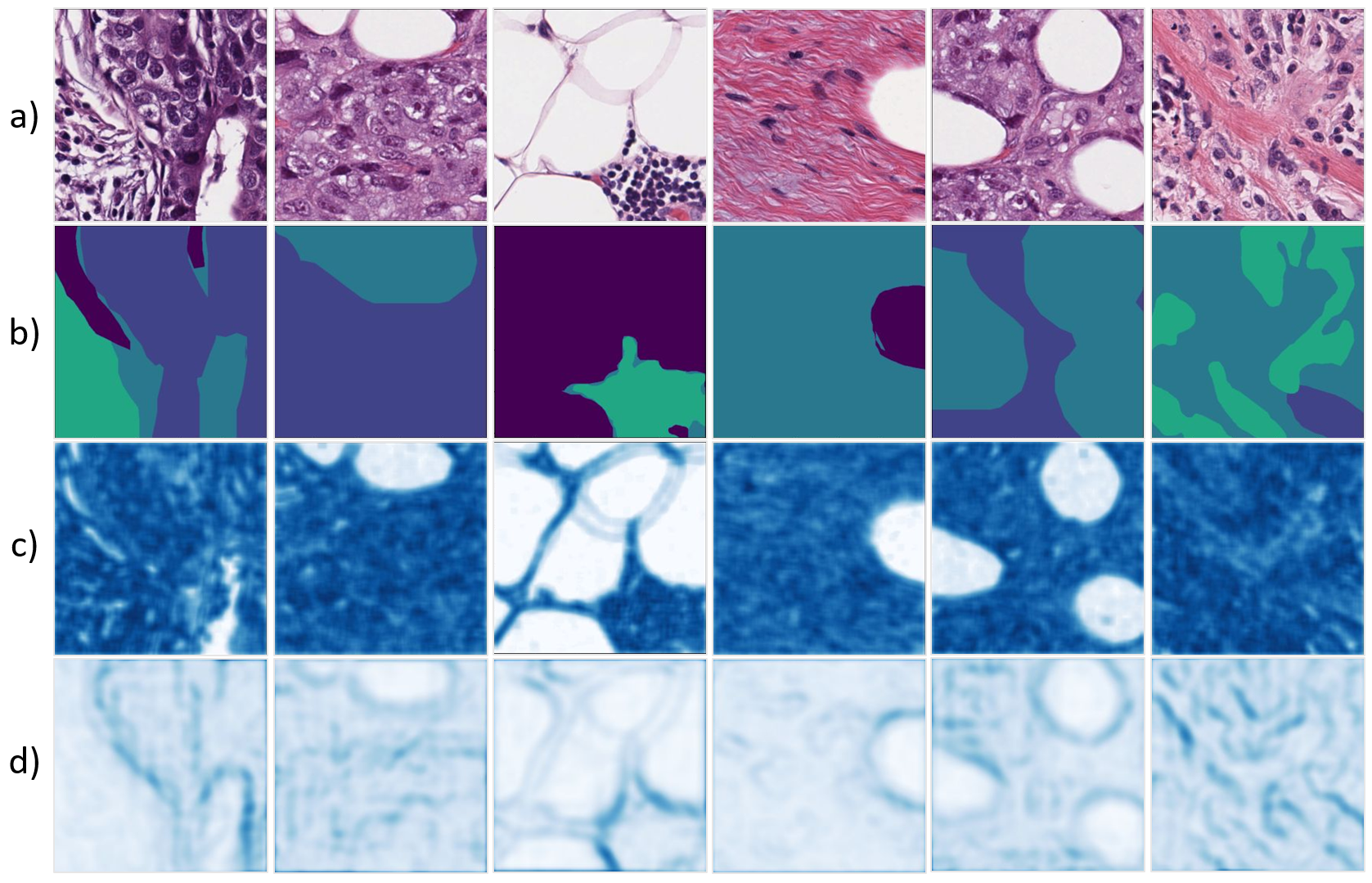}
    \captionsetup{width=0.99\textwidth}
    \caption{(a) Example images from BCSS test dataset. (b) Respective masks showing the foreground and background pixels. (c,d) Average euclidean distance $L^2$ between the central patch of size $21 \times 21$ with four overlapping patches in the immediate neighbours in RGB colour space and feature space respectively. Note that for feature space visualisation encoder embeddings were upsampled to map input size. The darker blue colour represents the low density regions corresponding to high average distance.}
    \label{fig:encoder_cluster_assumtion}
\end{figure}

Consistency regularisation based methods work on the basis of cluster assumption and have achieved SOTA results in semi-supervised classification and segmentation. The main idea behind consistency regularisation is to have high and low density regions where samples closer to each other are likely to share the same label forming a high density region with a low average distance. While the class boundaries are likely to be aligned with the low density regions i.e., high average distance. In order to observe cluster assumption, feature embeddings were extracted from CRCFP and were compared against RGB colour space as shown in Figure \ref{fig:encoder_cluster_assumtion}. Extracted feature maps were upsampled to match the size of the input image and then the average euclidean distance between each patch of size $21 \times 21$ centred around its 4 immediate spatial neighbours (left, right, top and bottom) was calculated.  It can be seen in Figure \ref{fig:encoder_cluster_assumtion}(d) that the class boundaries are much more aligned and apparent in the feature space as compared to the colour space where the boundaries doesn't align well thus violating cluster assumption. This can be due to the fact that the CNNs at higher layers tends to learn more semantic based features from the basic low-level features. Also, interestingly the background/fat represented in white colour in input images somewhat holds the high density regions because there is not much change in colour values for that region. While the rest of the tissue area is not very homogeneous in pixel values due to the presence of cells of various shapes and sizes.

\section{Conclusions}
In this work, we haved presented a novel consistency based semi-supervised learning based semantic segmentation framework for region and nuclei segmentation in histology images. The proposed method is invariant to varying contexts and perturbations making it efficient and robust for semantic segmentation tasks. We have shown that context-aware consistency learning can exploit unlabelled images efficiently with the help of cross-consistency training and entropy minimisation. Extensive experiments on two publicly available large histopathological datasets have shown the superiority of the CRCFP framework by achieving new SOTA results for semi-supervised semantic segmentation. Also, detailed ablation studies for different network parameters and components show the contribution of each network component, demonstrating the effectiveness of our method. Future directions include improvements to the proposed method with respect to improving the context-aware loss for minor classes and finding histology specific perturbation e.g., targeting stain variations, on a large multi-centric histopathological dataset. Large multi-centric data is vital for the validation of the study as the quality of downstream analysis is highly dependent on the segmented histology primitives. 

\section{Declaration of competing interest}
The authors declare that they have no known competing financial interests or personal relationships that could have appeared to influence the work reported in this paper.

\section{Acknowledgements}
RMSB is funded by the Chancellor Scholarship from University of Warwick. SEAR and NMR are part of the PathLAKE digital pathology consortium, which is funded by the Data to Early Diagnosis and Precision Medicine strand of the governments Industrial Strategy Challenge Fund, managed and delivered by UK Research and Innovation (UKRI). NMR was also supported by the UK Medical Research Council (grant award MR/P015476/1) and the Alan Turing Institute.

\bibliographystyle{unsrt}  
\bibliography{main}  

\begin{thebibliography}{10}

\bibitem{koohbanani2020nuclick}
Navid~Alemi Koohbanani, Mostafa Jahanifar, Neda~Zamani Tajadin, and Nasir
  Rajpoot.
\newblock Nuclick: a deep learning framework for interactive segmentation of
  microscopic images.
\newblock {\em Medical Image Analysis}, 65:101771, 2020.

\bibitem{shephard2021simultaneous}
Adam~J Shephard, Simon Graham, Saad Bashir, Mostafa Jahanifar, Hanya Mahmood,
  Ali Khurram, and Nasir~M Rajpoot.
\newblock Simultaneous nuclear instance and layer segmentation in oral
  epithelial dysplasia.
\newblock In {\em Proceedings of the IEEE/CVF International Conference on
  Computer Vision}, pages 552--561, 2021.

\bibitem{qaiser2019fast}
Talha Qaiser, Yee-Wah Tsang, Daiki Taniyama, Naoya Sakamoto, Kazuaki Nakane,
  David Epstein, and Nasir Rajpoot.
\newblock Fast and accurate tumor segmentation of histology images using
  persistent homology and deep convolutional features.
\newblock {\em Medical image analysis}, 55:1--14, 2019.

\bibitem{vu2021digital}
Quoc~Dang Vu, Caroline Fong, Katharina von Loga, Shan E~Ahmed Raza, Daniel
  Nava~Rodrigues, Bijal Patel, Clare Peckitt, Ruwaida Begum, Avani Athauda,
  Naureen Starling, et~al.
\newblock Digital histological markers based on routine h\&e slides to predict
  benefit from maintenance immunotherapy in esophagogastric adenocarcinoma.,
  2021.

\bibitem{jahanifar2021stain}
Mostafa Jahanifar, Adam Shepard, Neda Zamanitajeddin, RM~Bashir, Mohsin Bilal,
  Syed~Ali Khurram, Fayyaz Minhas, and Nasir Rajpoot.
\newblock Stain-robust mitotic figure detection for the mitosis domain
  generalization challenge.
\newblock In {\em International Conference on Medical Image Computing and
  Computer-Assisted Intervention}, pages 48--52. Springer, 2021.

\bibitem{lu2021data}
Ming~Y Lu, Drew~FK Williamson, Tiffany~Y Chen, Richard~J Chen, Matteo Barbieri,
  and Faisal Mahmood.
\newblock Data-efficient and weakly supervised computational pathology on
  whole-slide images.
\newblock {\em Nature biomedical engineering}, 5(6):555--570, 2021.

\bibitem{da2022digestpath}
Qian Da, Xiaodi Huang, Zhongyu Li, Yanfei Zuo, Chenbin Zhang, Jingxin Liu, Wen
  Chen, Jiahui Li, Dou Xu, Zhiqiang Hu, et~al.
\newblock Digestpath: a benchmark dataset with challenge review for the
  pathological detection and segmentation of digestive-system.
\newblock {\em Medical Image Analysis}, page 102485, 2022.

\bibitem{bilal2021development}
Mohsin Bilal, Shan E~Ahmed Raza, Ayesha Azam, Simon Graham, Mohammad Ilyas,
  Ian~A Cree, David Snead, Fayyaz Minhas, and Nasir~M Rajpoot.
\newblock Development and validation of a weakly supervised deep learning
  framework to predict the status of molecular pathways and key mutations in
  colorectal cancer from routine histology images: a retrospective study.
\newblock {\em The Lancet Digital Health}, 3(12):e763--e772, 2021.

\bibitem{echle2022artificial}
A~Echle, N~Ghaffari Laleh, P~Quirke, HI~Grabsch, HS~Muti, OL~Saldanha,
  SF~Brockmoeller, PA~van~den Brandt, GGA Hutchins, SD~Richman, et~al.
\newblock Artificial intelligence for detection of microsatellite instability
  in colorectal cancer—a multicentric analysis of a pre-screening tool for
  clinical application.
\newblock {\em ESMO open}, 7(2):100400, 2022.

\bibitem{shaban2022digital}
Muhammad Shaban, Shan E~Ahmed Raza, Mariam Hassan, Arif Jamshed, Sajid Mushtaq,
  Asif Loya, Nikolaos Batis, Jill Brooks, Paul Nankivell, Neil Sharma, et~al.
\newblock A digital score of tumour-associated stroma infiltrating lymphocytes
  predicts survival in head and neck squamous cell carcinoma.
\newblock {\em The Journal of Pathology}, 256(2):174--185, 2022.

\bibitem{mao2013stromal}
Yan Mao, Evan~T Keller, David~H Garfield, Kunwei Shen, and Jianhua Wang.
\newblock Stromal cells in tumor microenvironment and breast cancer.
\newblock {\em Cancer and Metastasis Reviews}, 32(1):303--315, 2013.

\bibitem{tabesh2007multifeature}
Ali Tabesh, Mikhail Teverovskiy, Ho-Yuen Pang, Vinay~P Kumar, David Verbel,
  Angeliki Kotsianti, and Olivier Saidi.
\newblock Multifeature prostate cancer diagnosis and gleason grading of
  histological images.
\newblock {\em IEEE transactions on medical imaging}, 26(10):1366--1378, 2007.

\bibitem{diamond2004use}
James Diamond, Neil~H Anderson, Peter~H Bartels, Rodolfo Montironi, and Peter~W
  Hamilton.
\newblock The use of morphological characteristics and texture analysis in the
  identification of tissue composition in prostatic neoplasia.
\newblock {\em Human pathology}, 35(9):1121--1131, 2004.

\bibitem{sirinukunwattana2015novel}
Korsuk Sirinukunwattana, David~RJ Snead, and Nasir~M Rajpoot.
\newblock A novel texture descriptor for detection of glandular structures in
  colon histology images.
\newblock In {\em Medical Imaging 2015: Digital Pathology}, volume 9420, pages
  186--194. SPIE, 2015.

\bibitem{anoraganingrum1999cell}
Dwi Anoraganingrum.
\newblock Cell segmentation with median filter and mathematical morphology
  operation.
\newblock In {\em Proceedings 10th International Conference on Image Analysis
  and Processing}, pages 1043--1046. IEEE, 1999.

\bibitem{ronneberger2015u}
Olaf Ronneberger, Philipp Fischer, and Thomas Brox.
\newblock U-net: Convolutional networks for biomedical image segmentation.
\newblock In {\em International Conference on Medical image computing and
  computer-assisted intervention}, pages 234--241. Springer, 2015.

\bibitem{Chen2017DeepLab}
Liang-Chieh Chen, George Papandreou, Iasonas Kokkinos, Kevin Murphy, and
  Alan~L. Yuille.
\newblock Deeplab: Semantic image segmentation with deep convolutional nets,
  atrous convolution, and fully connected crfs.
\newblock {\em arXiv:1606.00915 [cs]}, 5 2017.
\newblock arXiv: 1606.00915.

\bibitem{xie2021segformer}
Enze Xie, Wenhai Wang, Zhiding Yu, Anima Anandkumar, Jose~M Alvarez, and Ping
  Luo.
\newblock Segformer: Simple and efficient design for semantic segmentation with
  transformers.
\newblock {\em Advances in Neural Information Processing Systems},
  34:12077--12090, 2021.

\bibitem{zhang2022topformer}
Wenqiang Zhang, Zilong Huang, Guozhong Luo, Tao Chen, Xinggang Wang, Wenyu Liu,
  Gang Yu, and Chunhua Shen.
\newblock Topformer: Token pyramid transformer for mobile semantic
  segmentation.
\newblock In {\em Proceedings of the IEEE/CVF Conference on Computer Vision and
  Pattern Recognition}, pages 12083--12093, 2022.

\bibitem{lindman2019annotations}
Karin Lindman, Jer{\'o}mino~F Rose, Martin Lindvall, Claes Lundstrom, and
  Darren Treanor.
\newblock Annotations, ontologies, and whole slide images--development of an
  annotated ontology-driven whole slide image library of normal and abnormal
  human tissue.
\newblock {\em Journal of pathology informatics}, 10(1):22, 2019.

\bibitem{shaban2019novel}
Muhammad Shaban, Syed~Ali Khurram, Muhammad~Moazam Fraz, Najah Alsubaie, Iqra
  Masood, Sajid Mushtaq, Mariam Hassan, Asif Loya, and Nasir~M Rajpoot.
\newblock A novel digital score for abundance of tumour infiltrating
  lymphocytes predicts disease free survival in oral squamous cell carcinoma.
\newblock {\em Scientific reports}, 9(1):1--13, 2019.

\bibitem{kather2019deep}
Jakob~Nikolas Kather, Alexander~T Pearson, Niels Halama, Dirk J{\"a}ger,
  Jeremias Krause, Sven~H Loosen, Alexander Marx, Peter Boor, Frank Tacke,
  Ulf~Peter Neumann, et~al.
\newblock Deep learning can predict microsatellite instability directly from
  histology in gastrointestinal cancer.
\newblock {\em Nature medicine}, 25(7):1054--1056, 2019.

\bibitem{bejnordi2015multi}
Babak~Ehteshami Bejnordi, Geert Litjens, Meyke Hermsen, Nico Karssemeijer, and
  Jeroen~AWM van~der Laak.
\newblock A multi-scale superpixel classification approach to the detection of
  regions of interest in whole slide histopathology images.
\newblock In {\em Medical Imaging 2015: Digital Pathology}, volume 9420, pages
  99--104. SPIE, 2015.

\bibitem{bashir2022novel}
Raja Muhammad~Saad Bashir, Muhammad Shaban, Shan E~Ahmed Raza, Syed~Ali
  Khurram, and Nasir Rajpoot.
\newblock A novel framework for coarse-grained semantic segmentation of
  whole-slide images.
\newblock In {\em Annual Conference on Medical Image Understanding and
  Analysis}, pages 425--439. Springer, 2022.

\bibitem{jahanifar2021robust}
Mostafa Jahanifar, Neda~Zamani Tajeddin, Navid~Alemi Koohbanani, and Nasir~M
  Rajpoot.
\newblock Robust interactive semantic segmentation of pathology images with
  minimal user input.
\newblock In {\em Proceedings of the IEEE/CVF International Conference on
  Computer Vision}, pages 674--683, 2021.

\bibitem{long2015Fully}
Jonathan Long, Evan Shelhamer, and Trevor Darrell.
\newblock Fully convolutional networks for semantic segmentation.
\newblock {\em arXiv:1411.4038 [cs]}, 3 2015.
\newblock arXiv: 1411.4038.

\bibitem{zhao2017pyramid}
Hengshuang Zhao, Jianping Shi, Xiaojuan Qi, Xiaogang Wang, and Jiaya Jia.
\newblock Pyramid scene parsing network.
\newblock In {\em Proceedings of the IEEE conference on computer vision and
  pattern recognition}, pages 2881--2890, 2017.

\bibitem{sun2019deep}
Ke~Sun, Bin Xiao, Dong Liu, and Jingdong Wang.
\newblock Deep high-resolution representation learning for human pose
  estimation.
\newblock In {\em Proceedings of the IEEE/CVF conference on computer vision and
  pattern recognition}, pages 5693--5703, 2019.

\bibitem{vaswani2017attention}
Ashish Vaswani, Noam Shazeer, Niki Parmar, Jakob Uszkoreit, Llion Jones,
  Aidan~N Gomez, {\L}ukasz Kaiser, and Illia Polosukhin.
\newblock Attention is all you need.
\newblock {\em Advances in neural information processing systems}, 30, 2017.

\bibitem{zheng2021rethinking}
Sixiao Zheng, Jiachen Lu, Hengshuang Zhao, Xiatian Zhu, Zekun Luo, Yabiao Wang,
  Yanwei Fu, Jianfeng Feng, Tao Xiang, Philip~HS Torr, et~al.
\newblock Rethinking semantic segmentation from a sequence-to-sequence
  perspective with transformers.
\newblock In {\em Proceedings of the IEEE/CVF conference on computer vision and
  pattern recognition}, pages 6881--6890, 2021.

\bibitem{lee2013pseudo}
Dong-Hyun Lee et~al.
\newblock Pseudo-label: The simple and efficient semi-supervised learning
  method for deep neural networks.
\newblock In {\em Workshop on challenges in representation learning, ICML},
  volume~3, page 896, 2013.

\bibitem{zou2020pseudoseg}
Yuliang Zou, Zizhao Zhang, Han Zhang, Chun-Liang Li, Xiao Bian, Jia-Bin Huang,
  and Tomas Pfister.
\newblock Pseudoseg: Designing pseudo labels for semantic segmentation.
\newblock {\em arXiv preprint arXiv:2010.09713}, 2020.

\bibitem{chen2021semi}
Xiaokang Chen, Yuhui Yuan, Gang Zeng, and Jingdong Wang.
\newblock Semi-supervised semantic segmentation with cross pseudo supervision.
\newblock In {\em Proceedings of the IEEE/CVF Conference on Computer Vision and
  Pattern Recognition}, pages 2613--2622, 2021.

\bibitem{goodfellow2014yoshua}
Ian~J Goodfellow, Jean Pouget-Abadie, Mehdi Mirza, Bing Xu, David Warde-Farley,
  Sherjil Ozair, and Aaron Courville.
\newblock Yoshua bengio generative adversarial networks.
\newblock {\em arXiv preprint arXiv:1406.2661}, 2014.

\bibitem{badrinarayanan2015deep}
Vijay Badrinarayanan, Alex Kendall, and Roberto~Cipolla SegNet.
\newblock A deep convolutional encoder-decoder architecture for image
  segmentation.
\newblock {\em arXiv preprint arXiv:1511.00561}, 5, 2015.

\bibitem{kumar2017semi}
Abhishek Kumar, Prasanna Sattigeri, and Tom Fletcher.
\newblock Semi-supervised learning with gans: Manifold invariance with improved
  inference.
\newblock {\em Advances in neural information processing systems}, 30, 2017.

\bibitem{miyato2018virtual}
Takeru Miyato, Shin-ichi Maeda, Masanori Koyama, and Shin Ishii.
\newblock Virtual adversarial training: a regularization method for supervised
  and semi-supervised learning.
\newblock {\em IEEE transactions on pattern analysis and machine intelligence},
  41(8):1979--1993, 2018.

\bibitem{wei2018revisiting}
Yunchao Wei, Huaxin Xiao, Honghui Shi, Zequn Jie, Jiashi Feng, and Thomas~S
  Huang.
\newblock Revisiting dilated convolution: A simple approach for weakly-and
  semi-supervised semantic segmentation.
\newblock In {\em Proceedings of the IEEE conference on computer vision and
  pattern recognition}, pages 7268--7277, 2018.

\bibitem{ouali2020semi}
Yassine Ouali, C{\'e}line Hudelot, and Myriam Tami.
\newblock Semi-supervised semantic segmentation with cross-consistency
  training.
\newblock In {\em Proceedings of the IEEE/CVF Conference on Computer Vision and
  Pattern Recognition}, pages 12674--12684, 2020.

\bibitem{lai2021semi}
Xin Lai, Zhuotao Tian, Li~Jiang, Shu Liu, Hengshuang Zhao, Liwei Wang, and
  Jiaya Jia.
\newblock Semi-supervised semantic segmentation with directional context-aware
  consistency.
\newblock In {\em Proceedings of the IEEE/CVF Conference on Computer Vision and
  Pattern Recognition}, pages 1205--1214, 2021.

\bibitem{grandvalet2004semi}
Yves Grandvalet and Yoshua Bengio.
\newblock Semi-supervised learning by entropy minimization.
\newblock {\em Advances in neural information processing systems}, 17, 2004.

\bibitem{saito2019semi}
Kuniaki Saito, Donghyun Kim, Stan Sclaroff, Trevor Darrell, and Kate Saenko.
\newblock Semi-supervised domain adaptation via minimax entropy.
\newblock In {\em Proceedings of the IEEE/CVF International Conference on
  Computer Vision}, pages 8050--8058, 2019.

\bibitem{wu2022cross}
Huisi Wu, Zhaoze Wang, Youyi Song, Lin Yang, and Jing Qin.
\newblock Cross-patch dense contrastive learning for semi-supervised
  segmentation of cellular nuclei in histopathologic images.
\newblock In {\em Proceedings of the IEEE/CVF Conference on Computer Vision and
  Pattern Recognition}, pages 11666--11675, 2022.

\bibitem{amgad2019structured}
Mohamed Amgad, Habiba Elfandy, Hagar Hussein, Lamees~A Atteya, Mai~AT Elsebaie,
  Lamia~S Abo~Elnasr, Rokia~A Sakr, Hazem~SE Salem, Ahmed~F Ismail, Anas~M
  Saad, et~al.
\newblock Structured crowdsourcing enables convolutional segmentation of
  histology images.
\newblock {\em Bioinformatics}, 35(18):3461--3467, 2019.

\bibitem{kumar2019multi}
Neeraj Kumar, Ruchika Verma, Deepak Anand, Yanning Zhou, Omer~Fahri Onder,
  Efstratios Tsougenis, Hao Chen, Pheng-Ann Heng, Jiahui Li, Zhiqiang Hu,
  et~al.
\newblock A multi-organ nucleus segmentation challenge.
\newblock {\em IEEE transactions on medical imaging}, 39(5):1380--1391, 2019.

\bibitem{Badrinarayanan2016SegNet}
Vijay Badrinarayanan, Alex Kendall, and Roberto Cipolla.
\newblock Segnet: A deep convolutional encoder-decoder architecture for image
  segmentation.
\newblock {\em arXiv:1511.00561 [cs]}, 10 2016.
\newblock arXiv: 1511.00561.

\bibitem{wang2016deep}
Jiazhuo Wang, John~D MacKenzie, Rageshree Ramachandran, and Danny~Z Chen.
\newblock A deep learning approach for semantic segmentation in histology
  tissue images.
\newblock In {\em International Conference on Medical Image Computing and
  Computer-Assisted Intervention}, pages 176--184. Springer, 2016.

\bibitem{naylor2017nuclei}
Peter Naylor, Marick La{\'e}, Fabien Reyal, and Thomas Walter.
\newblock Nuclei segmentation in histopathology images using deep neural
  networks.
\newblock In {\em 2017 IEEE 14th international symposium on biomedical imaging
  (ISBI 2017)}, pages 933--936. IEEE, 2017.

\bibitem{yu2015multi}
Fisher Yu and Vladlen Koltun.
\newblock Multi-scale context aggregation by dilated convolutions.
\newblock {\em arXiv preprint arXiv:1511.07122}, 2015.

\bibitem{azam2020novel}
A~Azam, S~Bashir, SA~Khurram, D~Snead, and N~Rajpoot.
\newblock A novel deep learning-based diagnostic algorithm for detection and
  segmentation of amyloid in digital whole slide images.
\newblock In {\em VIRCHOWS ARCHIV}, volume 477, pages S214--S214. SPRINGER ONE
  NEW YORK PLAZA, SUITE 4600, NEW YORK, NY, UNITED STATES, 2020.

\bibitem{he2015spatial}
Kaiming He, Xiangyu Zhang, Shaoqing Ren, and Jian Sun.
\newblock Spatial pyramid pooling in deep convolutional networks for visual
  recognition.
\newblock {\em IEEE transactions on pattern analysis and machine intelligence},
  37(9):1904--1916, 2015.

\bibitem{graham2019mild}
Simon Graham, Hao Chen, Jevgenij Gamper, Qi~Dou, Pheng-Ann Heng, David Snead,
  Yee~Wah Tsang, and Nasir Rajpoot.
\newblock Mild-net: Minimal information loss dilated network for gland instance
  segmentation in colon histology images.
\newblock {\em Medical image analysis}, 52:199--211, 2019.

\bibitem{zhu2019asymmetric}
Zhen Zhu, Mengde Xu, Song Bai, Tengteng Huang, and Xiang Bai.
\newblock Asymmetric non-local neural networks for semantic segmentation.
\newblock In {\em Proceedings of the IEEE/CVF International Conference on
  Computer Vision}, pages 593--602, 2019.

\bibitem{huang2019ccnet}
Zilong Huang, Xinggang Wang, Lichao Huang, Chang Huang, Yunchao Wei, and Wenyu
  Liu.
\newblock Ccnet: Criss-cross attention for semantic segmentation.
\newblock In {\em Proceedings of the IEEE/CVF international conference on
  computer vision}, pages 603--612, 2019.

\bibitem{fraz2020fabnet}
MM~Fraz, Syed~Ali Khurram, Simon Graham, Muhammad Shaban, Mariam Hassan, Asif
  Loya, and Nasir~M Rajpoot.
\newblock Fabnet: feature attention-based network for simultaneous segmentation
  of microvessels and nerves in routine histology images of oral cancer.
\newblock {\em Neural Computing and Applications}, 32(14):9915--9928, 2020.

\bibitem{dosovitskiy2020image}
Alexey Dosovitskiy, Lucas Beyer, Alexander Kolesnikov, Dirk Weissenborn,
  Xiaohua Zhai, Thomas Unterthiner, Mostafa Dehghani, Matthias Minderer, Georg
  Heigold, Sylvain Gelly, et~al.
\newblock An image is worth 16x16 words: Transformers for image recognition at
  scale.
\newblock {\em arXiv preprint arXiv:2010.11929}, 2020.

\bibitem{karimi2022medical}
Davood Karimi, Haoran Dou, and Ali Gholipour.
\newblock Medical image segmentation using transformer networks.
\newblock {\em IEEE Access}, 10:29322--29332, 2022.

\bibitem{van2020survey}
Jesper~E Van~Engelen and Holger~H Hoos.
\newblock A survey on semi-supervised learning.
\newblock {\em Machine Learning}, 109(2):373--440, 2020.

\bibitem{wei2017object}
Yunchao Wei, Jiashi Feng, Xiaodan Liang, Ming-Ming Cheng, Yao Zhao, and
  Shuicheng Yan.
\newblock Object region mining with adversarial erasing: A simple
  classification to semantic segmentation approach.
\newblock In {\em Proceedings of the IEEE conference on computer vision and
  pattern recognition}, pages 1568--1576, 2017.

\bibitem{berthelot2019mixmatch}
David Berthelot, Nicholas Carlini, Ian Goodfellow, Nicolas Papernot, Avital
  Oliver, and Colin~A Raffel.
\newblock Mixmatch: A holistic approach to semi-supervised learning.
\newblock {\em Advances in neural information processing systems}, 32, 2019.

\bibitem{berthelot2019remixmatch}
David Berthelot, Nicholas Carlini, Ekin~D Cubuk, Alex Kurakin, Kihyuk Sohn, Han
  Zhang, and Colin Raffel.
\newblock Remixmatch: Semi-supervised learning with distribution alignment and
  augmentation anchoring.
\newblock {\em arXiv preprint arXiv:1911.09785}, 2019.

\bibitem{bashir2020hydramix}
Raja Muhammad~Saad Bashir, Talha Qaiser, Shan E~Ahmed Raza, and Nasir~M
  Rajpoot.
\newblock Hydramix-net: A deep multi-task semi-supervised learning approach for
  cell detection and classification.
\newblock In {\em Interpretable and Annotation-Efficient Learning for Medical
  Image Computing}, pages 164--171. Springer, 2020.

\bibitem{laine2016temporal}
Samuli Laine and Timo Aila.
\newblock Temporal ensembling for semi-supervised learning.
\newblock {\em arXiv preprint arXiv:1610.02242}, 2016.

\bibitem{tarvainen2017mean}
Antti Tarvainen and Harri Valpola.
\newblock Mean teachers are better role models: Weight-averaged consistency
  targets improve semi-supervised deep learning results.
\newblock {\em Advances in neural information processing systems}, 30, 2017.

\bibitem{chen2020simple}
Ting Chen, Simon Kornblith, Mohammad Norouzi, and Geoffrey Hinton.
\newblock A simple framework for contrastive learning of visual
  representations.
\newblock In {\em International conference on machine learning}, pages
  1597--1607. PMLR, 2020.

\bibitem{he2020momentum}
Kaiming He, Haoqi Fan, Yuxin Wu, Saining Xie, and Ross Girshick.
\newblock Momentum contrast for unsupervised visual representation learning.
\newblock In {\em Proceedings of the IEEE/CVF conference on computer vision and
  pattern recognition}, pages 9729--9738, 2020.

\bibitem{koohbanani2021self}
Navid~Alemi Koohbanani, Balagopal Unnikrishnan, Syed~Ali Khurram, Pavitra
  Krishnaswamy, and Nasir Rajpoot.
\newblock Self-path: Self-supervision for classification of pathology images
  with limited annotations.
\newblock {\em IEEE Transactions on Medical Imaging}, 40(10):2845--2856, 2021.

\bibitem{hadsell2006dimensionality}
Raia Hadsell, Sumit Chopra, and Yann LeCun.
\newblock Dimensionality reduction by learning an invariant mapping.
\newblock In {\em 2006 IEEE Computer Society Conference on Computer Vision and
  Pattern Recognition (CVPR'06)}, volume~2, pages 1735--1742. IEEE, 2006.

\bibitem{liu2021domain}
Weizhe Liu, David Ferstl, Samuel Schulter, Lukas Zebedin, Pascal Fua, and
  Christian Leistner.
\newblock Domain adaptation for semantic segmentation via patch-wise
  contrastive learning.
\newblock {\em arXiv preprint arXiv:2104.11056}, 2021.

\bibitem{chopra2005learning}
Sumit Chopra, Raia Hadsell, and Yann LeCun.
\newblock Learning a similarity metric discriminatively, with application to
  face verification.
\newblock In {\em 2005 IEEE Computer Society Conference on Computer Vision and
  Pattern Recognition (CVPR'05)}, volume~1, pages 539--546. IEEE, 2005.

\bibitem{schroff2015facenet}
Florian Schroff, Dmitry Kalenichenko, and James Philbin.
\newblock Facenet: A unified embedding for face recognition and clustering.
\newblock In {\em Proceedings of the IEEE conference on computer vision and
  pattern recognition}, pages 815--823, 2015.

\bibitem{sohn2016improved}
Kihyuk Sohn.
\newblock Improved deep metric learning with multi-class n-pair loss objective.
\newblock {\em Advances in neural information processing systems}, 29, 2016.

\bibitem{oord2018representation}
Aaron van~den Oord, Yazhe Li, and Oriol Vinyals.
\newblock Representation learning with contrastive predictive coding.
\newblock {\em arXiv preprint arXiv:1807.03748}, 2018.

\bibitem{grill2020bootstrap}
Jean-Bastien Grill, Florian Strub, Florent Altch{\'e}, Corentin Tallec, Pierre
  Richemond, Elena Buchatskaya, Carl Doersch, Bernardo Avila~Pires, Zhaohan
  Guo, Mohammad Gheshlaghi~Azar, et~al.
\newblock Bootstrap your own latent-a new approach to self-supervised learning.
\newblock {\em Advances in neural information processing systems},
  33:21271--21284, 2020.

\bibitem{french2019semi}
Geoff French, Samuli Laine, Timo Aila, Michal Mackiewicz, and Graham Finlayson.
\newblock Semi-supervised semantic segmentation needs strong, varied
  perturbations.
\newblock {\em arXiv preprint arXiv:1906.01916}, 2019.

\bibitem{ke2020guided}
Zhanghan Ke, Di~Qiu, Kaican Li, Qiong Yan, and Rynson~WH Lau.
\newblock Guided collaborative training for pixel-wise semi-supervised
  learning.
\newblock In {\em European conference on computer vision}, pages 429--445.
  Springer, 2020.

\bibitem{li2019signet}
Jiahui Li, Shuang Yang, Xiaodi Huang, Qian Da, Xiaoqun Yang, Zhiqiang Hu,
  Qi~Duan, Chaofu Wang, and Hongsheng Li.
\newblock Signet ring cell detection with a semi-supervised learning framework.
\newblock In {\em International conference on information processing in medical
  imaging}, pages 842--854. Springer, 2019.

\bibitem{lai2021semii}
Zhengfeng Lai, Chao Wang, Zin Hu, Brittany~N Dugger, Sen-Ching Cheung, and
  Chen-Nee Chuah.
\newblock A semi-supervised learning for segmentation of gigapixel
  histopathology images from brain tissues.
\newblock In {\em 2021 43rd Annual International Conference of the IEEE
  Engineering in Medicine \& Biology Society (EMBC)}, pages 1920--1923. IEEE,
  2021.

\bibitem{cheng2020self}
Hsien-Tzu Cheng, Chun-Fu Yeh, Po-Chen Kuo, Andy Wei, Keng-Chi Liu, Mong-Chi Ko,
  Kuan-Hua Chao, Yu-Ching Peng, and Tyng-Luh Liu.
\newblock Self-similarity student for partial label histopathology image
  segmentation.
\newblock In {\em European Conference on Computer Vision}, pages 117--132.
  Springer, 2020.

\bibitem{chen2019domain}
Minghao Chen, Hongyang Xue, and Deng Cai.
\newblock Domain adaptation for semantic segmentation with maximum squares
  loss.
\newblock In {\em Proceedings of the IEEE/CVF International Conference on
  Computer Vision}, pages 2090--2099, 2019.

\bibitem{weinstein2013cancer}
John~N Weinstein, Eric~A Collisson, Gordon~B Mills, Kenna~R Shaw, Brad~A
  Ozenberger, Kyle Ellrott, Ilya Shmulevich, Chris Sander, and Joshua~M Stuart.
\newblock The cancer genome atlas pan-cancer analysis project.
\newblock {\em Nature genetics}, 45(10):1113--1120, 2013.

\bibitem{He2016Deep}
Kaiming He, Xiangyu Zhang, Shaoqing Ren, and Jian Sun.
\newblock Deep residual learning for image recognition.
\newblock pages 770--778. 2016 IEEE Conference on Computer Vision and Pattern
  Recognition (CVPR), 6 2016.
\newblock ISSN: 1063-6919.

\bibitem{deng2009imagenet}
Jia Deng, Wei Dong, Richard Socher, Li-Jia Li, Kai Li, and Li~Fei-Fei.
\newblock Imagenet: A large-scale hierarchical image database.
\newblock In {\em 2009 IEEE conference on computer vision and pattern
  recognition}, pages 248--255. Ieee, 2009.

\bibitem{caron2020unsupervised}
Mathilde Caron, Ishan Misra, Julien Mairal, Priya Goyal, Piotr Bojanowski, and
  Armand Joulin.
\newblock Unsupervised learning of visual features by contrasting cluster
  assignments.
\newblock {\em Advances in Neural Information Processing Systems},
  33:9912--9924, 2020.

\bibitem{lu2018nuclear}
Cheng Lu, David Romo-Bucheli, Xiangxue Wang, Andrew Janowczyk, Shridar Ganesan,
  Hannah Gilmore, David Rimm, and Anant Madabhushi.
\newblock Nuclear shape and orientation features from h\&e images predict
  survival in early-stage estrogen receptor-positive breast cancers.
\newblock {\em Laboratory investigation}, 98(11):1438--1448, 2018.

\bibitem{kather2020pan}
Jakob~Nikolas Kather, Lara~R Heij, Heike~I Grabsch, Chiara Loeffler, Amelie
  Echle, Hannah~Sophie Muti, Jeremias Krause, Jan~M Niehues, Kai~AJ Sommer,
  Peter Bankhead, et~al.
\newblock Pan-cancer image-based detection of clinically actionable genetic
  alterations.
\newblock {\em Nature cancer}, 1(8):789--799, 2020.

\bibitem{diao2021human}
James~A Diao, Jason~K Wang, Wan~Fung Chui, Victoria Mountain, Sai~Chowdary
  Gullapally, Ramprakash Srinivasan, Richard~N Mitchell, Benjamin Glass, Sara
  Hoffman, Sudha~K Rao, et~al.
\newblock Human-interpretable image features derived from densely mapped cancer
  pathology slides predict diverse molecular phenotypes.
\newblock {\em Nature communications}, 12(1):1--15, 2021.

\bibitem{mcinnes2018umap}
Leland McInnes, John Healy, and James Melville.
\newblock Umap: Uniform manifold approximation and projection for dimension
  reduction.
\newblock {\em arXiv preprint arXiv:1802.03426}, 2018.

\end{thebibliography}

\end{document}